\documentclass[10pt,twocolumn,letterpaper]{article}

\usepackage{cvpr}              % To produce the CAMERA-READY version
\usepackage[accsupp]{axessibility}
\usepackage{graphicx}
\usepackage{amsmath}
\usepackage{amssymb}
\usepackage{mathtools}
\usepackage{booktabs}
\usepackage{pifont}
\usepackage{xcolor}
\usepackage{graphicx}
\usepackage{diagbox}
\usepackage{multirow}
\usepackage{bbm}
\newcommand{\cmark}{\ding{51}}%
\newcommand{\xmark}{\ding{55}}%

\usepackage{enumerate}
\usepackage[pagebackref,breaklinks,colorlinks]{hyperref}

\usepackage[capitalize]{cleveref}
\crefname{section}{Sec.}{Secs.}
\Crefname{section}{Section}{Sections}
\Crefname{table}{Table}{Tables}
\crefname{table}{Tab.}{Tabs.}

\def\cvprPaperID{6458} % *** Enter the CVPR Paper ID here
\def\confName{CVPR}
\def\confYear{2022}

\begin{document}

\title{Balanced Contrastive Learning for Long-Tailed Visual Recognition}
\author{Jianggang Zhu$^{1,2}$\thanks{Indicates equal contribution.}, Zheng Wang$^{1,2}$\footnotemark[1], Jingjing Chen$^{1,2}$\thanks{Jingjing-Chen is the corresponding author.}, Yi-Ping Phoebe Chen$^3$ and Yu-Gang Jiang$^{1,2}$\\
\textsuperscript{1}{Shanghai Key Lab of Intelligent Information Processing, School of Computer Science, Fudan University}\\
\textsuperscript{2}Shanghai Collaborative Innovation Center on Intelligent Visual Computing\\
\textsuperscript{3}{Department of Computer Science and Information Technology, La Trobe University}\\
{\tt\small \{jgzhu20,zhengwang17,chengjingjing,ygj\}@fudan.edu.cn, phoebe.chen@latrobe.edu.au}}

% \author{Jianggang Zhu, Zheng Wang\\
% Shanghai Key Lab of Intelligent Information Processing\\
% School of Computer Science, Fudan University\\
% {\tt\small \{jgzhu20,zhengwang17\}@fudan.edu.cn}
% \and
% }
% \author{Yi-Ping Phoebe Chen\\
% La Trobe University\\
% {\tt\small phoebe.chen@latrobe.edu.au}
% }
\maketitle

%%%%%%%%% ABSTRACT
\begin{abstract}
Real-world data typically follow a long-tailed distribution, where a few majority categories occupy most of the data while most minority categories contain a limited number of samples. Classification models minimizing cross-entropy struggle to represent and classify the tail classes. Although the problem of learning unbiased classifiers has been well studied, methods for representing imbalanced data are under-explored. In this paper, we focus on representation learning for imbalanced data. Recently, supervised contrastive learning has shown promising performance on balanced data recently. However, through our theoretical analysis, we find that for long-tailed data, it fails to form a regular simplex which is an ideal geometric configuration for representation learning. To correct the optimization behavior of SCL and further improve the performance of long-tailed visual recognition, we propose a novel loss for balanced contrastive learning (BCL). Compared with SCL, we have two improvements in BCL: class-averaging, which balances the gradient contribution of negative classes; class-complement, which allows all classes to appear in every mini-batch. The proposed balanced contrastive learning (BCL) method satisfies the condition of forming a regular simplex and assists the optimization of cross-entropy. Equipped with BCL, the proposed two-branch framework can obtain a stronger feature representation and achieve competitive performance on long-tailed benchmark datasets such as CIFAR-10-LT, CIFAR-100-LT, ImageNet-LT, and iNaturalist2018. Our code is available at \href{https://github.com/FlamieZhu/BCL}{this URL}.
\end{abstract}

%%%%%%%%% BODY TEXT
\begin{figure}[t]
    \centering
    \includegraphics[width=0.8\linewidth]{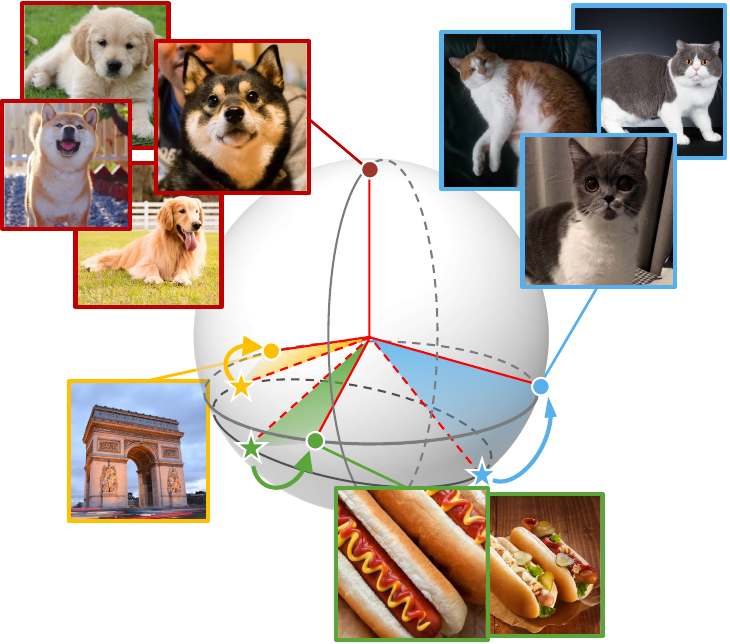}
    \caption{Illustration of Balanced Contrastive Learning. Head classes dominant the training procedure of SCL and compress the representation space of tail classes on the hypersphere (denoted by $\bigstar$). BCL learns an embedding space that treats all classes equally and forms a regular simplex (denoted by $\bullet$).}
    \label{fig:idea}
\end{figure}

\section{Introduction}
%What is long-tail and what is the problem
Deep neural networks have achieved remarkable success in a series of computer vision tasks, such as image recognition~\cite{alexnet,chen2020study,deng2019mixed}, video analysis~\cite{zhang2021token,su2020video}, object detection~\cite{fasterrcnn}, etc. These achievements are owing largely to the availability of large-scale dataset such as ImageNet~\cite{imagenet},
%benefits mostly from training with high quality large-scale datasets, i.e. ImageNet~\cite{imagenet}, 
 where each class has sufficient and equal amount of training samples. However, 
%where each class is carefully checked and has enough data for learning. 
real-world datasets are often imbalanced, where many classes have only a few samples and few classes have a great number of samples. Deep models trained with such unbalanced data usually generalize badly on balanced testing data, especially for rare classes. Improving recognition performance with unbalanced data poses a huge challenge to modern deep learning methods. 

To tackle the problem of learning with imbalanced data, early methods mainly focus on re-sampling the training data ~\cite{oversample_systematic,undersample,oversample_dynamic,oversample_effect} or re-weighting the loss functions ~\cite{CBLoss,reweight_learning,reweight_learningtotail} to pay more attention to rare classes.
%re-balanced training data such as re-sampling~\cite{oversample_systematic,undersample,oversample_dynamic,oversample_effect}, or adjusting the loss function, such as re-weighting~\cite{CBLoss,reweight_learning,reweight_learningtotail}. 
Recently, diverse methods have emerged. For example, Logit compensation methods~\cite{LDAM,equalization,logitadjustment} calibrate distribution between the training data and the test data. Decoupling~\cite{decoupling} adopts a two-stage training scheme where the classifier is re-balanced in the second stage. The work in~\cite{ride} has multiple distribution-aware experts for responding to samples of different class frequencies. Nevertheless, contrastive learning approaches are less explored before, not until contrastive learning ~\cite{kcl,hybrid,paco} are introduced. We attach great importance to representation learning because it's the most remarkable capability of deep models.

%Introduce contrastive learning
In this paper, we focus on using supervised contrastive learning (SCL)~\cite{scl} to assist representation learning.
%extended a popular method of self-supervised contrastive learning~\cite{simclr, contrastivemultiviewcoding} for supervised classification, 
%which has the instance from the same class as positives against the negatives from others within the batch and allows for multiple positives per anchor, 
Supervised contrastive loss has achieved better performance than supervised cross-entropy loss on large-scale classification problems. The work in \cite{dissecting} then has explained in detail the reason for the excellent performance of SCL on the balanced datasets. 
%Specifically, SCL as well as CE form a same geometric configuration when they attain their lower bound of loss respectively, that is, representations of each class collapse to the vertices of a regular simplex, which empirically confers important benefits, such as better generalization performance~\cite{prevalence}. 
%However, when long-tailed distributions arise, there seems to be some vexed issues.
Despite the great success, some recent work \cite{paco, kcl} indicate that high-frequency classes dominate SCL for representing imbalanced data, which results in unsatisfactory performance across all classes. 
%The work in \cite{paco} attributes poor performance to that high-frequency classes have a higher low bound of loss than low-frequency classes, and therefore high-frequency classes dominate the training procedure.
% {We further analysis what will such phenomenon causes?} \wz{However, the cause of SCL's failure on imbalanced data is not clear.}{}
%This is the most important basis of our motivation, and the subsequent analysis will revolve around this.
%However, there is no more profound study to reveal how the long-tailed distribution affects the representation learning. 
To analyze the optimizing behavior of SCL in learning the representations for long-tailed data, we depict the geometric arrangement of representations of training instances when the lower bound of loss is achieved. Specifically, we decouple the lower bound of loss by deriving two competing dynamics: an attraction term and a repulsion term as in~\cite{dissecting}. 
We reveal that the long-tailed distribution mainly affects the repulsion term.
%Give explanation of the problem of SCL
%From this perspective, we first further analyze why SCL fails to model and find that high-frequency classes will get a higher low bound than low-frequency classes as in \cite{paco}.
At the minimal supervised contrastive loss, the representations of classes of long-tailed data no longer attain a regular simplex configuration. In other words, when all instances with the same label collapse to points, these points are not equidistant from each other. A regular simplex configuration empirically confers important benefits, such as better generalization performance~\cite{prevalence}. Besides, it has been proved to be the target geometric configuration of SCL on balanced data~\cite{dissecting}, hence forming a regular simplex configuration will benefit recognition on long-tailed data~\cite{fang2021exploring}.
%, and parts ways with that of balanced data.
%Furthermore, the geometry of regular simplex is not available under long-tailed distributions. 

%Propose BCL 
Inspired by our analysis, we urge the model learning on imbalanced data to form a regular simplex, and propose a balanced contrastive learning (BCL) method (illustrated in Fig.~\ref{fig:idea}). We have two modifications in BCL that distinguish it from SCL. First, class-complement introduces the class-center embeddings, i.e., prototypes, as instances for comparison in every mini-batch. Second, class-averaging has the gradient contributions of all negatives of each class averaged for every mini-batch. 
%BCL drives data samples of the same class to collapse to points and these points form the vertices of a regular simplex, 
Through these two improvements, BCL ensures that the overall lower bound of the loss is a class-independent constant, and alleviates the imbalance problem of SCL when representing long-tailed data.
%Our goal is to improve the prediction accuracy by learning better representations.
%Inspired by the aforementioned, we present a balanced contrastive loss to improve representation learning of long-tailed visual tasks, 
Furthermore, We adopt a cross-entropy loss with logit compensation to obtain a balanced classifier. Logit compensation can effectively alleviate overlooking tail classes in the classifier learning~\cite{LDAM,balancedsoftmax,logitadjustment,disentangling}. Overall, we propose a two-branch framework to implement the mentioned techniques, i.e., a contrastive learning branch with BCL and a classification branch with logit compensated cross-entropy.

%\wz{}{Recall the recent success of mixture-of-experts in long-tailed visual recognition, these ensemble-based methods are committed to getting better prediction results by routing diverse experts \cite{ride,TADE}. However, ensemble-based method avoids the problem of how to learn better representations from long-tailed datasets by dynamically weighting the output logits of each expert.
%Our goal is to improve the prediction accuracy by learning better representations. Through this way, we can also fine-tune the trained model for other downstream tasks.}
Our main contributions are as follows:
\begin{itemize}
    \item We present a theoretical analysis showing that supervised contrastive learning forms an undesired asymmetric geometry configuration for long-tailed data due to the overwhelming numerical dominance of the head classes.
    \item Motivated by our analysis, we extend supervised contrastive learning to balanced contrastive learning, which overcomes the imbalance problem and remains a regular simplex configuration of long-tailed data.
    \item The proposed two-branch framework combines the classification module and the balanced contrastive learning module, achieving competitive results on several popular long-tailed datasets.
\end{itemize}

\section{Related Work}
\noindent\textbf{Long-tailed Recognition}
Early solutions to address the long-tailed problem comprise two main ideas: re-sampling and re-weighting. 
Re-sampling methods undersample~\cite{oversample_systematic,undersample} high-frequency classes or oversample~\cite{oversample_dynamic,oversample_effect,oversample_systematic} low-frequency classes.
%So low-frequency classes are concentrated more.
Re-weighting methods~\cite{CBLoss,reweight_learning,reweight_learningtotail} assign different losses to different training samples for each class or each example.
%is achieved by providing less weight to high-frequency classes and more weight to low-frequency classes. Whereas this straightforward re-balancing method improves learning of the tail classes, it leads to underfitting or overfitting. 
%In addition, recently, a advanced re-balancing learning method improving long-tailed learning different from the above has emerged, known as logit compensation. 
%Ride~\cite{ride} has multiple distribution-awarded experts for responding to samples of different class frequency. 
%\noindent\textbf{\wz{One/Two-stage imbalance learning}{}}
Both BBN\cite{bbn} and Decoupling\cite{decoupling} indicate that the re-balancing method is detrimental to representation learning.
%, despite its ability to improve classifier learning. 
BBN dynamically adjusts the weights between features from the instance-balanced sampling branch and the reversed sampling branch.
%with an end-to-end learning framework. 
While Decoupling proposes a two-stage learning strategy that firstly obtains a good feature extractor and secondly fixes the feature extractor and fine-tunes the classifier.
Recently proposed logit compensation methods~\cite{LDAM,equalization,logitadjustment,decoupling} learn relatively larger margins between different classes based on the prior of class frequencies. For example, logit adjustment\cite{logitadjustment} derives the general form of the compensation value based on the optimal Bayesian classifier.
Our proposed framework simultaneously improves the representation learning with BCL and strengthens the classifier learning with logit compensation in an end-to-end manner.

\noindent\textbf{Supervised Contrastive Learning}
Contrastive learning (CL) trains the model in a pairwise way by aggregating semantically similar samples while excluding semantically dissimilar ones, which has been employed for feature representation learning in varies tasks~\cite{wang2022cross, wang2021visual, han2020self}. SimCLR\cite{simclr} and MoCo\cite{moco} are two typical types of self-supervised contrastive learning.
%SimCLR learns representations through a symmetric network, i.e., siamese network, which requires a large batch size. MoCo is able to utilize a substantial number of negative samples even with a small batch size by using the momentum encoder. 
%\cite{scl} extends a supervised form of contrastive learning by leveraging label information and consistently outperforms cross-entropy. 
%CL enhances the ability for self-supervised representation learning\cite{simclr,moco,swav}. 
SCL~\cite{scl} leverages label information for fully-supervised representation learning, leading to state-of-the-art performance for image classification. 
% SCL still employs the two-stage training. In the first stage, a sufficient feature extractor is learned. Then the feature extractor is fixed and a linear classifier is learned in the second stage.
%Generally,  both  self-supervised contrastive learning and supervised contrastive learning  adopt  two-stage  training  strategies. In the first stage, one can utilize contrastive loss to train a feature extractor, and train a linear classifier by cross-entropy in the second stage.
%\cite{scl} explored the infoNCE loss of supervised form, which builds on the contrastive self supervised literature by leveraging label information.

\noindent\textbf{Contrastive Learning for Long-tailed Recognition} Trained on long-tailed data, conventional contrastive learning can pose potential problems. SSP~\cite{rethinking} boosts long-tailed learning with self-supervised and semi-supervised contrastive learning. Hybrid-SC~\cite{hybrid} designs a two-branch network, using a supervised contrastive learning branch for learning better representations and a classifier branch for eliminating the bias of the classifier towards head classes. Despite our framework shares a similar two-branch architecture with Hybrid-SC, our framework differs from Hybrid-SC as we introduce BCL loss in the SCL branch for dealing with the domination of majority classes problem. Hybrid-PSC~\cite{hybrid} is proposed to overcome the memory bottleneck problem of SCL with a prototype for each class for contrast. PaCo~\cite{paco} overcomes the performance degradation of SCL by introducing a set of class-wise learnable centers. KCL~\cite{kcl} adopts the two-stage learning paradigm and uses the same number of positives for all classes in every batch. %Our proposed framework is most similar to Hybrid-SC~\cite{hybrid}, while we introduce BCL loss in the SCL branch for dealing with the domination of majority classes problem.
%\blue{To our knowledge, the recently proposed TSC~\cite{li2021targeted} is the most relevant work to ours, which urges the features of classes within a batch closer to the target features on the vertices of a regular simplex. TSC computes the targets without knowing the class semantics, while BCL uses the class prototypes optimized with logit compensation as targets.}
The recently proposed TSC~\cite{li2021targeted} is the most relevant work to ours, which urges the features of classes closer to the target features on the vertices of a regular simplex. Targets in TSC are learned without class semantics, while our BCL uses the class prototypes as extra samples.

\section{Method}
%First, we discuss the drawbacks of supervised contrastive loss for long-tailed visual recognition, then we show how the problem is solved by keeping a regular simplex configuration under the proposed balanced contrastive loss. Further, we present the overall framework equipped with balanced contrastive learning and the final optimization target.
%In this section, we will compare the differences between our proposed balanced supervised contrastive learning and the conventional supervised contrastive learning in detail. And we will explain the advantages of our method from different perspectives.

\subsection{Preliminaries}
In the image classification task, we aim to learn a complex function $\varphi$ mapping from an input space $\mathcal{X}$ to the target space $\mathcal{Y}=[K]=\{1,2,\dots,K\}$. The function $\varphi$ is usually implemented as the composition of an encoder $f:\mathcal{X}\to \mathcal{Z}\in\mathbb{R}^{h}$ and a linear classifier $W:\mathcal{Z}\to \mathcal{Y}$.
%and the $\mathrm{argmax}$ operation. 
The final classification accuracy strongly depends on the quality of the representations $\mathcal{Z}$. Therefore, we aim to learn a good encoder $f$ to improve long-tailed learning. Further, we have the following definitions to facilitate later analysis:

%\noindent \textbf{\textit{Definition 3.1 (Supervised contrastive loss)}}\ 
\noindent\textbf{Supervised contrastive loss.}
For an instance $x_i$ of representation $z_i$ in a batch $B$, supervised contrastive loss has the following expression:
\begin{equation}
    \mathcal{L}_{i}=-\frac{1}{\vert B_{y}\vert-1}\sum_{p\in B_{y}\setminus\lbrace{i}\rbrace}\log\frac{\exp(z_{i}\cdot z_{p}/\tau)}{\sum\limits_{k\in B\setminus\{i\}}\exp(z_{i}\cdot z_{k}/\tau)}
\end{equation}
where $B_{y}$ is a subset of $B$ that contains all samples of class $y$, and we further define $B^{C}_{y}$ as the complement set of $B_{y}$. $\vert \cdot\vert$ stands for the number of samples in the set. $\tau>0$ is a scalar temperature hyper parameter that controls tolerance to similar samples, and a small temperature tends to be less tolerant to similar samples \cite{understandingssl}. Note that we omit $\tau$ in the following contrastive losses for simplicity.

%\noindent \textbf{\textit{Definition 3.2 (Class-specific batch-wise loss)}} \ 
%\noindent\textbf{Class-specific batch-wise loss.}
Similar to \cite{dissecting}, we also introduce the class-specific batch-wise loss:
\begin{equation}
    \mathcal{L}_{SCL}(Z;Y,B,y) = \begin{cases}
    \sum_{i\in B_{y}} \mathcal{L}_{i} & if \  \vert B_{y}\vert > 1\\
    0 & else\\
    \end{cases}
\end{equation}

%\noindent \textbf{\textit{Definition 3.3 (Regular simplex)}} 
\noindent\textbf{Regular simplex.}
A set of points $\zeta_{1},\dots,\zeta_{K}\in \mathbb{R}^{h}$ form the vertices of a regular simplex inscribed in the hypersphere of radius $\rho>0$, if and only if the following conditions hold:
\begin{enumerate}[(1)]
    \item $\sum\nolimits_{i\in[K]}\zeta_{i}=0$
    \item $\Vert \zeta_{i} \Vert=\rho, \text{for} \ i \in {[K]}$
    \item $\exists d \in \mathbb{R}: d = \langle\zeta_{i},\zeta_{j}\rangle\ \quad \text{for}\ 1\leq i < j \leq K$
\end{enumerate}
where h, K $\in \mathbb{N}$ with $K \leq h+1$, and $\langle \cdot \rangle$ stands for the inner product operation. Regular simplex has a highly symmetric structure that all vertices are equally spaced.
%For convenience we default to $K \leq h+1$. 
Given a balanced dataset, it's worth mentioning that when supervised contrastive loss attain the minimum, representations of each class collapse to the vertices of a regular simplex spontaneously~\cite{prevalence, dissecting}. See the illustration in Fig.~\ref{fig:simplex}(a). 
%encourages the anchor and all positives to gather together and push away from all negatives, and when it, 
%We use the same definition of regular simplex as in~\cite{dissecting}.

\subsection{Analysis}\label{scetion:analysis}
%We start by revealing the drawback of supervised contrastive learning for long-tailed data, then we propose our solution in balanced contrastive learning and give a detailed analysis.
\begin{figure}[t] 
  \begin{subfigure}[t]{0.34\linewidth}
    \centering
    \includegraphics[scale=0.1275]{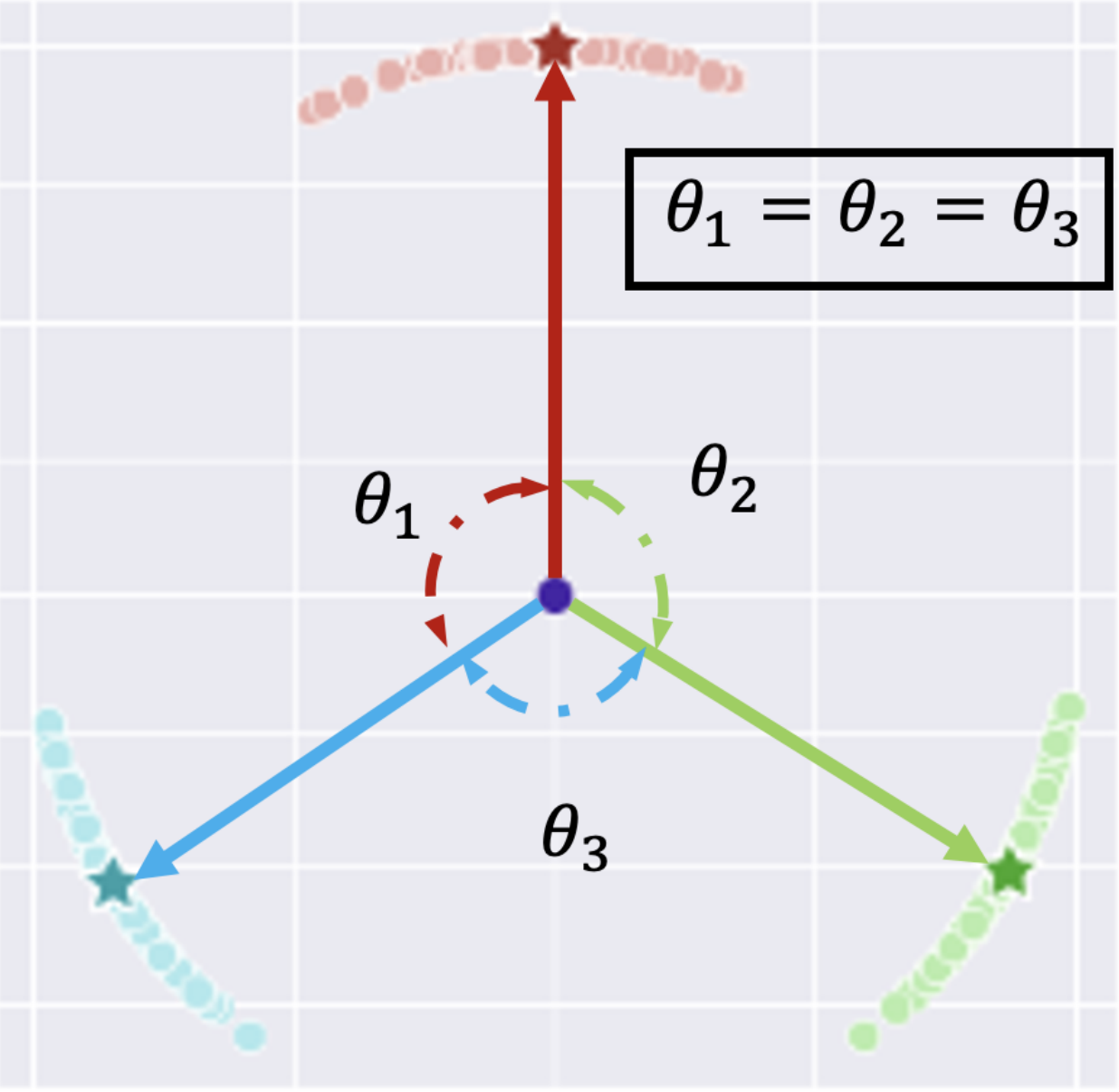}
    \label{1a} 
    \caption{}
  \end{subfigure}%%
  \begin{subfigure}[t]{0.34\linewidth}
    \centering
    \includegraphics[scale=0.1275]{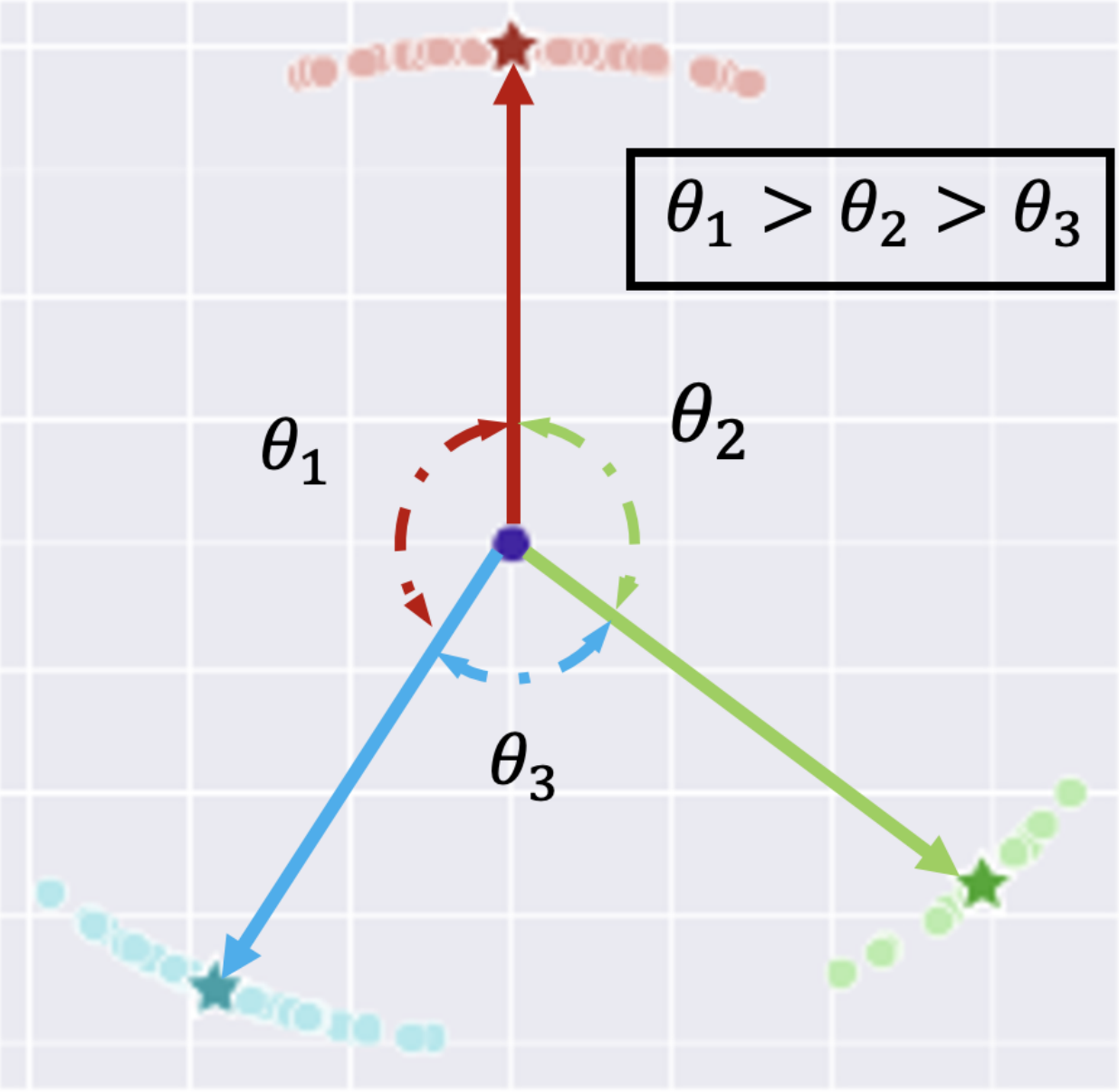} 
    \label{1b} 
    \caption{}
  \end{subfigure}%%
  \begin{subfigure}[t]{0.34\linewidth}
    \centering
    \includegraphics[scale=0.1275]{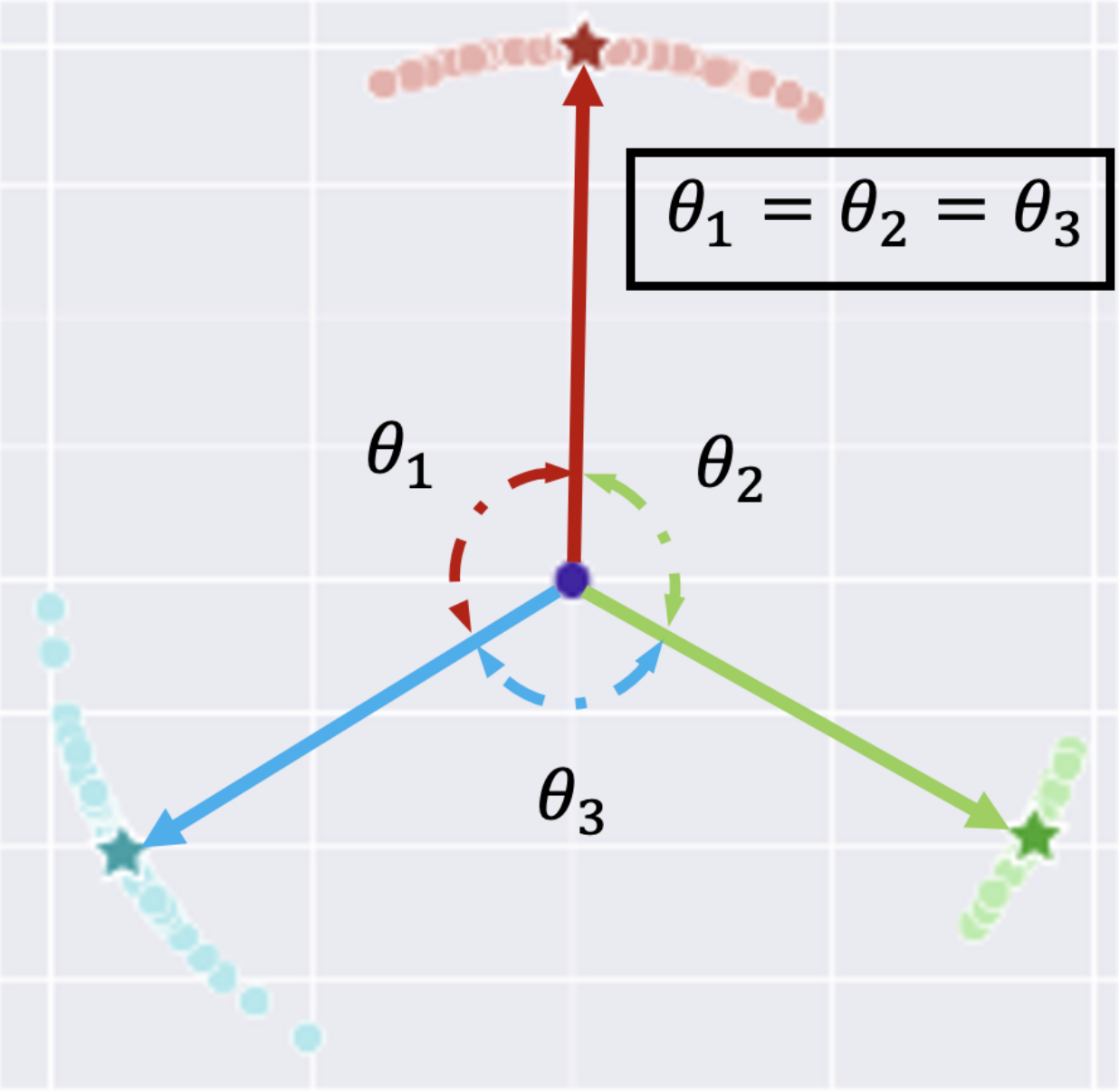} 
    \label{1c}
    \caption{}
    \end{subfigure}
    \vspace{-5pt}
    \caption{Illustration of the geometry configuration of data points with regard to (a)SCL on balanced data, (b) SCL on long-tailed data, and (c) BCL on long-tailed data in a plane.
    The $\bigstar$ represents the class-mean of each class. Different colors represent different classes.
    SCL enlarges the distances between high-frequency classes and reduces the distances between low-frequency classes as in (b), causing an asymmetrical geometry configuration of long-tailed data.
    }
    \label{fig:simplex}
\end{figure}
\noindent\textbf{Drawbacks of SCL.}\quad To clearly show the optimization behaviour of SCL on long-tailed data, we mainly focus on the variation of the geometry configuration formed by the representations of each class. 
%To analyze the optimization behaviour of SCL on long-tailed data,  when  its lower bound of loss is attained. %\wz{}{Previous work \cite{paco} has pointed out that high-frequency classes have a higher lower bound of loss and low-frequency classes have a lower one.} In other words, supervised contrastive loss concentrates more on head than tail.
Although representations of each class collapse to the vertices of a regular simplex when supervised contrastive loss attain its minimum on a balanced dataset, SCL forms an asymmetrical configuration on long-tailed data as shown in Fig.~\ref{fig:simplex}(b). In the following, we will give an in-depth analysis on the loss function to show why the geometry configuration changes for imbalanced data. In particular, we analysis the lower-bound of the loss. Since directly computing the lower bound on the whole long-tailed dataset is often intractable, we pay attention to the loss of a specific mini-batch instead.
% \textit{The equality is attained if and only if:}
% \begin{flalign*}
% &(1) \exists C_{i}(B,y), such\ that\ \forall j \in B_{y} \setminus{\lbrace i \rbrace}\ all\ inner\ products\\
% &\langle z_{i},z_{j} \rangle = C_{i}(B,y)\ are\ equal. \\
% &(2) \exists D_{i}(B,y), such\ that\ \forall k \in B_{y}^{C}\ all\ inner\ products\\
% &\langle z_{i},z_{k} \rangle = D_{i}(B,y)\ are\ equal. 
% \end{flalign*}
\begin{figure*}[t]
    \centering
    \includegraphics[width=0.9\linewidth]{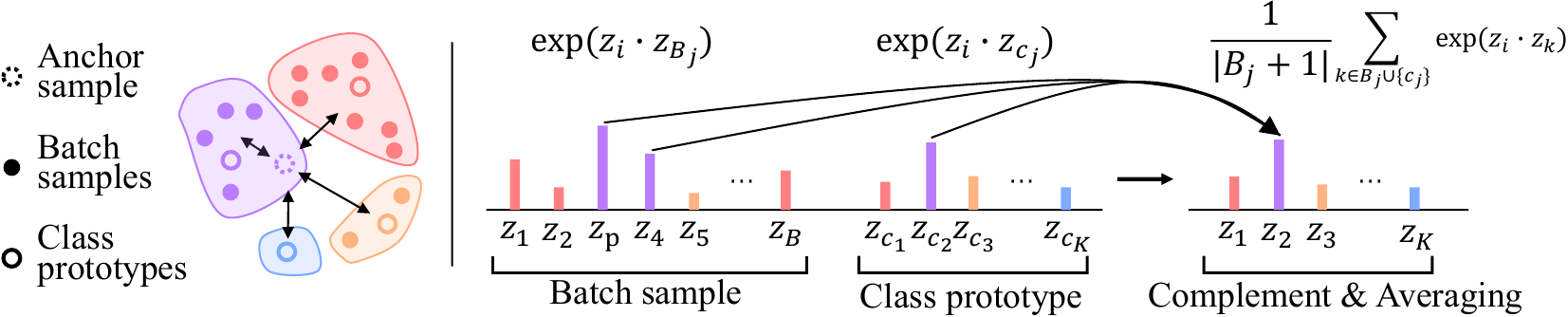}
    \caption{\textbf{Left: }Contrasting an anchor sample with others. 
    %The anchor sample should be compared with samples of all classes.
    \textbf{Right: }Applying both class-averaging and class-complement for a specific class. We average the similarity between the anchor sample and the batch sample as well as the class prototype. Note that the class colored blue does not appear in the mini-batch, so the similarity between the anchor and its prototype can be directly taken as the result.}
    \label{fig:average}
\end{figure*}

\noindent \textit{\textbf{Theorem 1.}} \textit{Assuming the normalization function is applied for feature embedding, let $Z=(z_{1},\dots,z_{N})\in \mathcal{Z}^{N}$ be an N point configuration with labels $Y=(y_{1},\dots
,y_{N})\in [K]^{N}$, where
$\mathcal{Z}=\lbrace z\in \mathbb{R}^{h}: \Vert z\Vert =1\rbrace$. The class-specific batch-wise loss is bounded by }
\begin{equation}
\small{
\begin{aligned}
    &\mathcal{L}_{SCL}(Z;Y,B,y) \geq \sum_{i\in B_{y}} \log(
    (\vert B_{y}\vert -1)+ \\\vspace{-10pt}
    &\vert B_{y}^{C}\vert
    \exp(\underbrace{
    \frac{1}{\vert B_{y}^{C}\vert} \sum_{k\in B_{y}^{C}} z_{i}\cdot z_{k}}_{repulsion\ term}\underbrace{-
    \frac{1}{\vert B_{y}\vert -1} \sum_{j\in B_{y}\setminus{\lbrace i \rbrace}} z_{i}\cdot z_{j}}_{attraction\ term}))
\end{aligned}
}
\end{equation}
\textit{{Proof.}}\ \textit{See Lemma S1 in~\cite{dissecting}.}

The above lower bound of SCL loss is derived by~\cite{dissecting}, which consists of a repulsion term and an attraction term. The attraction term leads to variability collapse\cite{prevalence} as training progresses, and all the within-class representations collapse to their class means in the end.
The attraction term only relates to samples within a specific class. It means that whether the dataset is balanced or not, samples within the same class should be as close as possible.

The attraction term leads to intra-class feature collapse regardless of the class frequency. While the repulsion term affects inter-class uniformity and is dominated by classes with higher frequency, thereby features in SCL are less separable.
We indicate that data imbalance mainly affects the repulsion term.
Obviously, the repulsion term is strongly related to the data distribution of the classes appeared within a mini-batch. When the dataset is long-tailed, almost every mini-batch we sampled is long-tailed.
This leads to the dominance of the head classes in the repulsion term and makes each sample farther away from the heads. However, due to the number of samples in each class being different, the distance between the head classes will be larger compared to the others. Additionally, for each sample, the gradients from negative head classes will be much larger than negative tail classes.
% \wz{Head classes contribution most of the gradient in the repulsion term.}{}
This unavoidably causes the loss to focus more on optimizing the head classes and leads to an asymmetrical geometry as shown in Fig.~\ref{fig:simplex}(b).

\begin{figure*}[]
    \centering
    \includegraphics[width=0.9\linewidth]{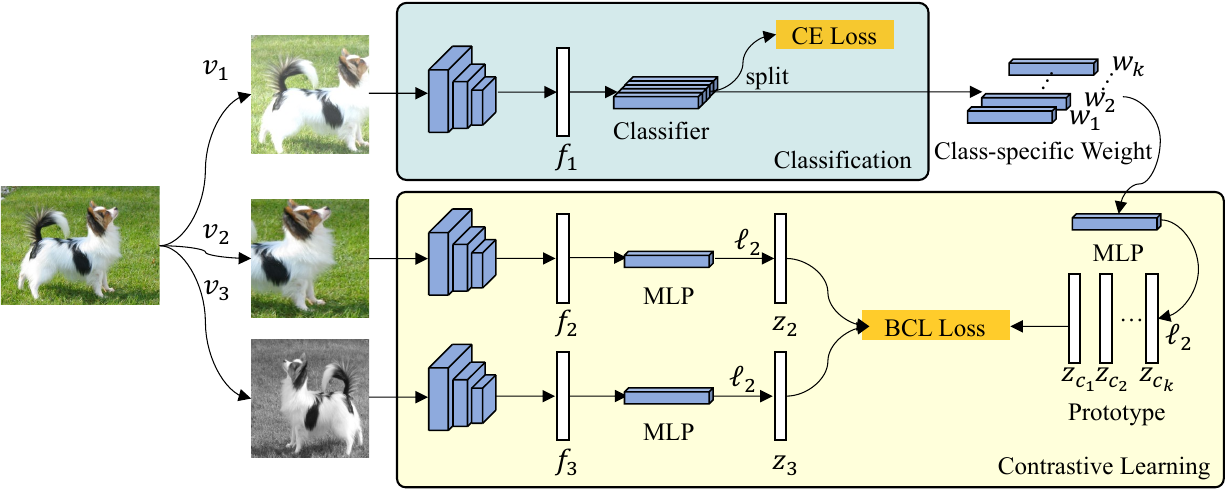}
    \caption{Overview of the proposed framework. The framework consists of a classification branch and a balanced contrastive learning branch. $v_2$ and $v_3$ adopt the same augmentation method different from $v_1$. The backbone is shared between two branches. The classifier weights are separately transformed by a MLP to be used as prototypes. All representations are $\ell_{2}$-normalized for balanced contrastive loss.}
    \label{fig:framework}
\end{figure*}
\noindent\textbf{Solution.}\quad We modify the supervised contrastive loss with two straightforward ideas, i.e., \textit{class-averaging} and \textit{class-complement}. The modified loss will lead to a regular simplex configuration of long-tailed data as we will show below.
% In order to avoid excessive concentration on head classes, \wz{a good intuition is to equilibrate class-specific batch-wise loss so that the lower bound is class-independent.}{} This can be achieved by averaging the contribution of each class in the loss for each sample. In the following, we abbreviate this operation as \textbf{\textit{class-averaging}}. Below, we give the lower bound after performing class-averaging.
To avoid excessive concentration on head classes, an intuitive approach is to equilibrate the gradients contributed by different negative classes. We call this operation as \textbf{\textit{class-averaging}}. The gradients from the negative samples of head classes are reduced. Below, we give the lower bound after performing class-averaging.

\noindent \textit{\textbf{Theorem 2.}} \textit{Let $Z,Y$ be defined as in Theorem 1, $\mathcal{Y}_{B} \subseteq [K]$ denotes the set of classes that appear within batch. the class-specific batch-wise loss after performing class-averaging is bounded by }
\begin{equation}
\small{
\begin{aligned}
    &\mathcal{L}_{BCL}(Z;Y,B,y) \geq \sum_{i\in B_{y}}\log (
    1+(\vert \mathcal{Y}_{B}\vert\! -\! 1) \times \exp(\\
    &\underbrace{
    \frac{1}{\vert \mathcal{Y}_{B}\vert \!-\! 1} \sum_{q\in\mathcal{Y}_{B}\setminus{\lbrace y \rbrace}} \frac{1}{\vert B_{q}\vert}
    \sum_{k \in B_{q} }
    z_{i}\!\cdot\! z_{k}}_{repulsion\ term}
    \underbrace{\!-
     \frac{1}{\vert B_{y}\vert \!-\! 1} \sum_{j\in B_{y}\setminus{\lbrace i \rbrace}} z_{i}\!\cdot\! z_{j}}_{attraction\ term}))
\end{aligned}
}
\end{equation}
\textit{{Proof.}} \ \textit{See the Supplementary Material. }

Consequently, head classes no longer dominate the repulsion term. Since each class is not sampled with equal probability, this may still lead to an unstable optimization and fail to form a regular simplex. To address this problem, we make all classes appear in every mini-batch and name this operation as \textbf{\textit{class-complement}}. Below, we give the overall lower bound after performing class-complement.

\noindent \textit{\textbf{Theorem 3.}} \textit{Let $Z,Y$ be defined as in Theorem 1,
if we have $\mathcal{Y}_{B} = \mathcal{Y}$ for every $B$, the overall loss is given by}
\begin{equation}
\label{eq:lower_bound_bcl}
    \mathcal{L}_{BCL}(Z;Y) \geq \vert \mathcal{D}\vert\log (
    1+( K -1) \exp(- \frac{K}{ K -1} ) )
\end{equation}
\textit{where $\mathcal{D}$ represents the dataset. Here, the normalization term is neglected. Recall that $Z$ is an N point configuration with labels $Y$, the equality of Eq.~\ref{eq:lower_bound_bcl} is attained if and only if the following conditions hold. There are $\zeta_{1},\dots,\zeta_{K} \in \mathbb{R}^{h}$ such that:}
\vspace{-5pt}
\begin{enumerate}[(1)]
    \item $\forall n \in [N]:  z_{n}=\zeta_{y_{n}}$
    \vspace{-5pt}
    \item $\zeta_{1},\dots,\zeta_{K} \ form\  a\  regular\ simplex$
\end{enumerate}
\vspace{-5pt}
\textit{{Proof.}} \ \textit{See the Supplementary Material. }

Note that condition (1) implies variability collapse, and condition (2) demonstrates the regular simplex structure.
When balanced contrastive loss attains its lower bound, each negative class contributes to the gradient equally. Additionally, BCL ensures that the loss of each sample is consistent and class-independent when attaining its lower bound, which implies that the learning will be less biased towards the head classes (Fig.~\ref{fig:simplex}(c)).
%\wz{}{Further, for a specific batch, BCL enables the lower bound to be consistent for each sample.}

\subsection{Balanced Contrastive Learning}\label{section:bcl}
%Now, we describe how the balanced contrastive learning works in practice with class-averaging and class-complement.
%Conventional supervised contrastive loss will converge when each class collapses to a point and these points form a regular simplex. When working on long-tailed datasets, this symmetrical structure will be broken as training progresses. Disappointingly, this leads to a imbalanced feature space, where head classes dominated most of the area. 
\noindent\textbf{Class-averaging}\quad The key idea is to average the instances of each class in a mini-batch so that each class has an approximate contribution for optimizing. Intuitively, it reduces the proportion of head classes in the denominator and emphasizes the importance of tails. In Section~\ref{scetion:analysis}, we take the loss form as $\mathcal{L}_{1}$ for analysis. However, there are other ways to implement class-averaging. Here, we give three loss forms as follows:
\begin{equation}
   \label{eq:l1}
    \mathcal{L}_{1}\!=\!-\frac{1}{\vert B_{y}\vert \!-\!1}\!\sum_{p\in B_{y}\!\setminus\!\{{i}\}}\!\log\!\frac{\exp(z_{i}\!\cdot\! z_{p})}{\!\sum\limits_{j\in \mathcal{Y}_{B}}\!\frac{1}{\vert B_{j}\vert}\!\sum\limits_{k\in B_{j}}\!\exp(z_{i}\!\cdot\!z_{k})}\\
\end{equation}
\vspace{-15pt}
\begin{equation}
\label{eq:l2}
    \mathcal{L}_{2}\!=\!-\frac{1}{\vert B_{y}\vert \!-1\!}\!\sum_{p\in B_{y}\!\setminus\!\{{i}\}}\!\log\!\frac{\exp(z_{i}\!\cdot\! z_{p})}{\!\sum\limits_{j\in \mathcal{Y}_{B}}\!\exp(\frac{1}{\vert B_{j}\vert}\!\sum\limits_{k\in B_{j}}\!z_{i}\!\cdot\!z_{k})}\\
\end{equation}
\vspace{-10pt}
\begin{equation}
\label{eq:l3}
   \mathcal{L}_{3}=-\log\frac{\exp(z_{i}\cdot z_{c_{y}})}{\sum\limits_{j\in 
   \mathcal{Y}} \exp(z_{i}\cdot z_{c_{j}})}
\end{equation}
where the $\vert B_{j} \vert$ term will minus one when the positive class is averaged.
%defined as $\vert B_{y}\vert \coloneqq \vert B_{y}\vert-1$.
The only difference between our $\mathcal{L}_{1}$ and $\mathcal{L}_{2}$ is that the averaging operation takes place in different positions. $\mathcal{L}_{1}$ performs averaging outside the exponential function, while $\mathcal{L}_{2}$ has averaging inside the exponential function. Since -$\log$ and $\exp$ are convex functions, which implies that  $\mathcal{L}_{1}\ge\mathcal{L}_{2}$ by Jensen's inequality. $\mathcal{L}_{3}$ take the form proposed in other prototype-based contrastive learning methods~\cite{dro-lt,hybrid,PCL}, where each sample is pulled towards its class prototype and pushed away from others. Note that $z\in\mathcal{Z}\subseteq\mathbb{R}^{h}$ is $\ell_2$-normalized for the inner product, so that $\Vert z \Vert_{2}=1$. 
%\wz{}{However, we cannot guarantee that the norm of the calculated prototype is equal to 1, even if re-normalizing the prototypes,it will lead to additional errors.} 
A comparison of their performance is in the experiment section. Below, we choose $\mathcal{L}_1$ for optimization.

 \noindent\textbf{Class-complement}\quad To have all classes appear in every mini-batch, we introduce class-center representations, i.e., prototypes for balanced contrastive learning.
%, which forces the lower bound of loss is a constant value for each sample.
Now we have the formulation of balanced contrastive loss as follows:
\begin{equation}
\label{eq:BCL}
\begin{aligned}
    &\mathcal{L}_{BCL}=-\frac{1}{\vert B_{y}\vert} \times \\
    &\sum_{p\in \lbrace B_{y}\setminus \lbrace i \rbrace \rbrace\cup \lbrace c_{y} \rbrace
    }
    \log\frac{\exp(z_{i}\!\cdot\! z_{p})}{
    \sum\limits_{j\in \mathcal{Y}}\frac{1}{\vert B_{j}\vert+1}
    \sum\limits_{k\in B_{j}\cup \lbrace c_{j} \rbrace}\exp(z_{i}\!\cdot\! z_{k})}
\end{aligned}
\end{equation}
%Balanced contrastive learning (BCL) ensures that all mini-batches have the same lower bound by introducing class-specific prototypes.
where $c_j$ is the index of the prototype.
In practice, we perform a nonlinear mapping of the classifier weights and regard the output as the prototype of each class. 

By applying both class-averaging and class-complement, the lower bound is a class-independent constant, avoiding the model's preference for head classes.
Note that in practice, we perform class-complement before class-averaging, as formulated in Eq.~\ref{eq:BCL} and illustrated in Fig.~\ref{fig:average}.

\noindent\textbf{Framework}\quad The overview of the proposed framework is shown in Fig.~\ref{fig:framework}. It consists of two main components: a classification branch and a contrastive learning branch. Both branches are trained simultaneously and share the same feature extractor. BCL is a unified end-to-end model, which is different from conventional contrastive learning methods that follow a two-stage training strategy. We have different augmentation methods for the two branches. Three different views are generated in total, where $v_{1}$ is the view used for the classification task, $v_{2}$ and $v_{3}$ are the pairwise views for the contrastive learning task.
%Multiple views of the data play a significant role in contrastive learning. %\wz{}{and recent research\cite{notbecontrastive,multiview} shows that the performance of downstream tasks usually rely heavily on the selection of the views we choose. From this principle, we adopt different data augmentation strategies for different downstream tasks.} 
% \wz{}{We will describe the details of these views in the experiment section and the ablation experiment will fully demonstrate the benefits of utilizing this data augmentation strategy.}
Following the work in~\cite{simclr,scl}, we utilize a symmetric architecture for the contrastive learning branch. We use a MLP with one hidden layer to obtain the representation $z_i$ for contrastive learning where $z_i=W_1\sigma(W_2 f_i)$ and $\sigma$ is a ReLU function. 
Instead of using mean embeddings~\cite{dro-lt} or learnable parameters~\cite{paco,hybrid} as prototypes, we are motivated by that the weights of the linear classifier are co-linear with these simplex vertices to which the classes collapse~\cite{dissecting,prevalence}. 
% The prototypes introduced for class-complement are adapted from classifier weights. 
Specifically, we have class-specific weights ${w_1, w_2, \dots, w_K}$ after a nonlinear transformation MLP as prototypes ${z_{c_{1}}, z_{c_{2}}, \dots,z_{c_{K}}}$.
Ablations of using different configurations of views and different forms of prototypes are given in the Supplementary Material.
%Specifically, in the classification branch, the classifier weights are separated to learn class-specific prototypes. 
All representations for contrastive learning is $\ell_{2}$ normalized to ensure the feature space is a unit hypersphere.

%\wz{}{As with the data augmentation strategy, the tradeoffs between classes in BCL are equally important.}
%\wz{The final output logit is the inner product of the sample embedding $f_{1,j}$ and the classifier weights $w_{j}$ plus the bias term $b_{j}$,}{} assuming that the norm of the embeddings are consistent, the logit value is attain its maximum value when the embedding and the corresponding class weights are linearly correlated. 
%Note that if one obtain the optimal lower bound, the embeddings of each class are linearly correlated with the corresponding classifier weights respectively. 
%Thus, in BCL, we regard the weights of the classifier after a simple nonlinear transformation as the prototype of each class. 
\noindent\textbf{Optimization with Logit Compensation}\quad
For long-tailed learning tasks, due to the imbalance of data, the output logit of the last classification layer usually exists bias. Logit compensation aims to eliminate the bias caused by the imbalance of data and learn the rectification of the boundary~\cite{logitadjustment,LDAM,equalization}.
The compensation can be applied during either training or testing. Previous work\cite{logitadjustment,LDAM,equalization,CBLoss,disentangling} illustrates the desirability of logit compensation in long-tailed visual tasks and it can be summarized as the following form
\begin{equation}
    \mathcal{L}_{LC}(y,\varphi(x))=-\alpha_{y}\log\frac{\exp (\varphi_{y}(x)+\delta_{y})}{\sum\limits_{y' \in \mathcal{[Y]}} \exp(\varphi_{y'}(x)+\delta_{y'})} 
\end{equation}
Here, $\alpha_{y}$ is the factor that controls the importance of class $y$, $\delta_{y}$ is the compensation for class $y$ and its value is related to class-frequency. We define $\alpha_{y}=1$, $\delta_{y}=\log \mathbb{P}_{y}$ as in \cite{logitadjustment} and perform logit compensation along with training, where $\mathbb{P}_{y}$ denotes the class prior of label $y$.

Finally, we have the following loss for training:
\begin{equation}
    \mathcal{L}=\lambda \mathcal{L}_{LC} + \mu \mathcal{L}_{BCL}
\end{equation}
where $\lambda$ and $\mu$ are hyperparameters that control the impact of $\mathcal{L}_{LC}$ and $\mathcal{L}_{BCL}$, respectively.
In addition, the contrastive branch only intends for the backbone to learn the desired feature embeddings.
%, and it does not involve the testing phase for prediction.

\section{Experiment}
%In this section, we first introduce four popular long-tailed image classification datasets and then present our results on these datasets. After this, we describe our proposed BCL implementation details and compare it with the state-of-the-art long-tail visual recognition method.
\subsection{Dataset}
%To validate the effectiveness of our proposed BCL, we conduct various experiments on four representative long-tailed image datasets.
\noindent\textbf{Long-Tailed CIFAR-10 and CIFAR-100}\quad CIFAR-10-LT and CIFAR-100-LT are the subsets of CIFAR-10 and CIFAR-100, respectively. Both CIFAR-10 and CIFAR-100 contain 50,000 images for training and 10,000 images for the validation of size 32 $\times$ 32 with 10 and 100 classes respectively. Following \cite{bbn,LDAM,CBLoss}, we use the same long-tailed version for a fair comparison. The imbalanced factor $\beta$ is defined by $\beta=N_{max}/N_{min}$, and this reflects the degree of imbalance in the data. The imbalance factors used in the experiment are set to 100, 50, and 10.

\noindent\textbf{ImageNet-LT}\quad ImageNet-LT is proposed in \cite{largescale}, which is a long-tailed version of vanilla ImageNet by sampling a subset following the Pareto distribution with power value $\alpha=0.6$. It consists of 115.8K images of 1000 classes in total with 1280 to 5 images per class. 

\noindent\textbf{iNaturalist 2018}\quad iNaturalist 2018~\cite{inaturalist} is a large-scale dataset containing 437.5K images from 8,142 classes. It is long-tailed by nature with an extremely imbalanced distribution. In addition to long-tailed recognition, this dataset is also used for evaluating the fine-grained classification task.

% \begin{minipage}{0.5\textwidth}
%  \begin{minipage}[t]{0.45\textwidth}
%   \centering
%     \makeatletter\def\@captype{table}\makeatother
%     \centering
%     \begin{tabular}{c c}
%     \hline
%          Methods & Top-1 Acc.  \\
%          \hline
%         $\mathcal{L}_{1}$ & 51.9\\
%         $\mathcal{L}_{2}$ & 50.2\\
%         $\mathcal{L}_{3}$ & 51.0\\
%         \hline
%     \end{tabular}
%     \caption{Comparison between different class-averaging methods on CIFAR-100 with an imbalance factor of 100.}
%   \end{minipage}
%   \begin{minipage}[t]{0.45\textwidth}
%   \centering
%     \makeatletter\def\@captype{table}\makeatother
%     \centering
%     \begin{tabular}{c c}
%     \hline
%          Methods & Top-1 Acc.  \\
%          \hline
%         Class-specific & 53.9\\
%         Learnable & 53.3\\
%         EMA & 53.3\\
%         \hline
%     \end{tabular}
%     \caption{Comparison between different class-averaging methods on CIFAR-100 with an imbalance factor of 100.  }
%   \end{minipage}
% \end{minipage}

\begin{table}[t]
    \centering
    \small
    \begin{tabular}{c c}
    \hline
         Methods & Top-1 Acc.  \\
         \hline
        $\mathcal{L}_{1}$ & 51.9\\
        $\mathcal{L}_{2}$ & 50.2\\
        $\mathcal{L}_{3}$ & 51.0\\
        \hline
\end{tabular}
    \caption{Ablation study for different class-averaging methods. All models run for 200 epochs with the same training scheme.}
    %$\mathcal{L}_{1}$ and $\mathcal{L}_{2}$ are intuitive ways to average the proportion of each class in the loss and $\mathcal{L}_{3}$ is also known as prototypical contrastive loss.} 
    \label{tab:comparison of scl}
\end{table}

% \begin{table}[t]
%     \centering
%     \begin{tabular}{c c}
%     \hline
%          Methods & Top-1 Acc.  \\
%          \hline
%         Class-specific & 53.9\\
%         Learnable & 53.3\\
%         EMA & 53.3\\
%         \hline
% \end{tabular}
%     \caption{Comparison between different ways of computing prototypes. We run the model on CIFAR-100 with an imbalance factor of 100 for 400 epochs.}
%     \label{tab:comparison of center}
% \end{table}

\begin{table}[t]
    \centering
    \small
    \begin{tabular}{c c c c c}
    \hline
         LC & SC  & Complement & Averaging & Top-1 Acc.  \\
         \hline
        \cmark & \xmark &\xmark & \xmark & 50.8\\
        \cmark & \cmark &\xmark & \xmark & 52.4 \\
        \cmark & \cmark &\cmark & \xmark & 52.3 \\
        \cmark & \cmark &\xmark & \cmark & 52.0 \\
        \cmark & \cmark &\cmark & \cmark & 53.9\\
        \hline
\end{tabular}
    \caption{Ablation study for the primary components of BCL. LC and SC denote logit compensation and supervised contrastive loss. Complement and averaging stand for class-complement and class-averaging, respectively. All models run for 400 epochs.}
    \label{ablation}
\end{table}

\begin{table*}[t]
    \centering
    \small
    \begin{tabular}{c||ccc|ccc}
     \hline
     Method & \multicolumn{3}{c|}{CIFAR-100-LT} & \multicolumn{3}{c}{CIFAR-10-LT}\\ 
     \hline
     Imbalance Factor & 100 & 50 & 10 & 100 & 50 & 10\\
     \hline
    %  CE & 38.32 & 43.85 & 55.71 & 70.36 & 74.81 & 86.39 \\
    %  \hline
     SSP\cite{rethinking}& 43.43 & 47.11 & 58.91 & 77.83 & 82.13 & 88.53\\
     Focal loss$^\dagger$ \cite{focal} & 38.41 & 44.32 & 55.78 & 70.38 & 76.72 & 86.66\\
     CB-Focal \cite{cbfocal} & 39.60 & 45.17 & 57.99 & 74.57 & 79.27 & 87.10\\
     BBN\cite{bbn} & 42.56 & 47.02 & 59.12 & 79.82 & 81.18 & 88.32\\
     Casual model\cite{casual} & 44.10 & 50.30 & 59.60 & 80.60 & 83.60 & 88.50\\
     LDAM-DRW\cite{LDAM} & 42.04 & 46.62 & 58.71 & 77.03 & 81.03 & 88.16\\
     ResLT\cite{reslt}& 48.21 & 52.71 & 62.01 &82.40 &85.17& 89.70\\
     Hybrid-SC\cite{hybrid} & 46.72& 51.87&63.05&81.40 &85.36&\textbf{91.12}\\
     MetaSAug-LDAM\cite{metasaug} & 48.01 & 52.27 & 61.28 & 80.66 & 84.34 & 89.68 \\
     \hline
     BCL(ours) & \textbf{51.93} & \textbf{56.59} & \textbf{64.87} & \textbf{84.32} & \textbf{87.24} & \textbf{91.12} \\
     \hline
\end{tabular}
    \caption{Top-1 accuracy of ResNet-32 on CIFAR-100-LT and CIFAR-10-LT. The best results are marked in bold. $\dagger$ denotes results borrowed from \cite{hybrid}. We report the results of 200 epochs.}
    \label{tab:cifar100-top1}
\end{table*}

\subsection{Implementation details}
For both CIFAR-10-LT and CIFAR-100-LT, we use the ResNet-32 as the backbone. 
Same to~\cite{paco}, we use AutoAugment~\cite{autoaugment} and Cutout~\cite{cutout} as data augmentation strategies for the classification branch and SimAugment~\cite{simclr} for the contrastive learning branch. To control the impact of $\mathcal{L}_{LC}$ and $\mathcal{L}_{BCL}$, $\lambda$ is set to 2.0, $\mu$ is 0.6, and the temperature $\tau$ is set to 0.1. We set the batch size as 256 and the weight decay as 5e$-$4.
The dimension of the hidden layer and the output layer of MLP are set to 512 and 128, respectively.
We run BCL for 200 epochs with the learning rate warms up to 0.15 within the first 5 epochs and decays at epoch 160 and 180 with a step size of 0.1.
Following \cite{paco}, we also run the model for 400 epochs, where the learning rate warms up to 0.15 within the first 10 epochs and decays at epoch 360 and 380 with a step size of 0.1. 
We train the above models with one Nvidia GeForce 1080Ti GPU. 

For ImageNet-LT, we use ResNet-50\cite{resnet} and ResNeXt-50-32x4d\cite{resnext} as our backbone. We run BCL for 90 epochs with an initial learning rate of 0.1 and the weight decay is 5e$-$4.
For iNaturalist, we use ResNet-50 as our backbone and run BCL for 100 epochs using an initial learning rate of 0.2 and the weight decay is 1e$-$4.
For both ImageNet-LT and iNaturalist 2018, we use cosine scheduling for learning rate, $\lambda$ is set to 1.0 and $\mu$ is set to 0.35. The batch size is set to 256. 
%\blue{We use the same data augmentation strategy as in \cite{paco}.} 
We use the RandAug augmentation strategy for the classification branch, and SimAug for the contrastive learning branch. The performances of different augmentation strategies are in the Supplementary Material.
To reduce memory consumption, the dimension of the output layer of MLP is set to 1024 for both datasets. We use the cosine classifier. All models are trained using SGD optimizer with a momentum set to 0.9.
For a fair comparison, we reproduce PaCo of ResNext-50 on ImageNet-LT for 180 epochs. %We train the above models with 4 Nvidia GeForce 3090 GPUs.
\begin{table}[t]
    \centering
    \small
   \begin{tabular}{l||cccc}
    \hline
         Methods & Many & Medium & Few & All  \\
         \hline
        200 epochs \\ 
        % CE(baseline) & 65.5 & 37.9 & 7.4 & 38.3\\
        $\tau$-norm$^\dagger$\cite{decoupling} & 61.4 & 42.5 & 15.7 & 41.4\\
        Hybrid-SC\cite{hybrid} & - & - & - & 46.7\\
        MetaSAug-LDAM\cite{metasaug} & - & - & - & 48.0\\
        DRO-LT\cite{dro-lt} & 64.7 & 50.0 & 23.8 & 47.3\\
        RIDE(3 experts)\cite{ride} & \textbf{68.1} & 49.2 & 23.9 & 48.0\\
        \hline
        BCL(Ours) & 67.2 & \textbf{53.1} & \textbf{32.9} & \textbf{51.9}\\
         \hline
        400 epochs \\ 
        Balanced Softmax$^\ddagger$\cite{balancedsoftmax} & - & - & - & 50.8\\
        PaCo\cite{paco} & - & - & - & 52.0\\
        \hline
        BCL(Ours) & \textbf{69.7} & \textbf{53.8} & \textbf{35.5} & \textbf{53.9}\\
        \hline
\end{tabular}
    \caption{Top-1 accuracy of ResNet-32 on CIFAR-100-LT with an imbalance factor of 100. We report the results of 200 epochs and 400 epochs. $\dagger$ and $\ddagger$ denote results borrowed from \cite{dro-lt} and \cite{paco}.}
    \label{tab:cifar100-100}
\end{table}

\subsection{Ablation study}
We perform several ablation studies to characterize the proposed BCL method. All experiments are performed on CIFAR-100 with an imbalance factor of 100. First, we compare the performances of different class-averaging implementations (i.e., $\mathcal{L}_{1}$, $\mathcal{L}_{2}$ and $\mathcal{L}_{3}$) mentioned in Section~\ref{section:bcl}.
%the overall accuracies of Eqs.~\ref{eq:l1}, \ref{eq:l2}, and \ref{eq:l3} in Section~\ref{section:bcl} is compared in Table~\ref{tab:comparison of scl}. 
The main difference between $\mathcal{L}_{1}$ and $\mathcal{L}_{2}$ is the order in which the averaging operations are implemented. For $\mathcal{L}_{3}$, we use the prototype implemented in our work instead of the average of all embeddings of the same class. As shown in Table \ref{tab:comparison of scl}, $\mathcal{L}_{1}$ achieves the best performance, which is consistent with our previous analysis. Surprisingly, $\mathcal{L}_{3}$ achieves better performance than $\mathcal{L}_{2}$, which may be attributed to the well-represented characteristics of prototypes.

To demonstrate the superiority of the balanced contrastive loss, we compare the performance of the primary components of the loss in Table \ref{ablation}. We use the cross-entropy loss with logit compensation (LC) as the vanilla baseline. %and additionally report the performance based on this.
% Compared with Balanced Softmax in Table \ref{tab:cifar100-100}, we find there is no difference in accuracy between Balanced Softmax and LC.
SC denotes a baseline that adds the contrastive learning branch with conventional supervised contrastive loss. Class-complement and class-averaging are the main techniques of the proposed balanced contrastive loss. We show that using either class-complement or class-averaging alone cannot improve the overall accuracy. In contrast, a significant performance boost can be obtained when both of them are applied, which indicates both components are indispensable components to achieve stronger performance.
%forming the regular simplex for long-tailed data and improving the results.

% \begin{table}[t]
%     \centering
%   \begin{tabular}{l|| c c c c}
%     \hline
%     Methods & Many & Medium & Few & All \\
%          \hline
%         $\tau$-norm & 56.6 & 44.2 & 27.4& 46.7\\
%         LWS &  57.1 & 45.2 & 29.3&47.7\\
%         DisAlign  & 61.3 & 52.2 & 31.4 & 52.9\\
%         RIDE(2 experts) & - & - & - & 54.4\\
%         Paco* & - & - & - & 57.0\\
%         BCL(Ours) & 66.2 & 53.5 & 36.4 &56.0\\
%         \hline
% \end{tabular}
%     \caption{Top-1 accuracy of ResNet-50 on ImageNet-LT. For a fair comparison, we report results of 90 epochs. *Paco runs for 400 epoches.}
%     \label{tab:imagenet-lt-top1-resnet}
% \end{table}

% \subsection{Comparison to state-of-the-art methods}
% RIDE~\cite{ride} trains model from an ensemble-based manner and achieves quite excellent performance on several long-tailed benchmarks. RIDE obtains high performance by routing diverse experts, however, this is different from our motivation of boosting long-tailed learning via training a sufficient encoder and a balanced feature space.
% PaCo~\cite{paco} uses parametric centers for contrastive learning which can be regarded as a classifier. PaCo distinguishes between the center and samples by weighting, while the prototypes in BCL supplements the samples of a class so that all classes within the batch appear.
% In addition, BCL ensures that the loss of each sample is consistent and class-independent when the lower bound is attained, which implies that the learning will not be biased towards the head classes.

\subsection{Main results}
\noindent \textbf{Long-tailed CIFAR}\quad
The comparison results between the proposed BCL and other existing methods on long-tailed CIFAR are shown in Table \ref{tab:cifar100-top1}. As can be seen from the table, BCL consistently outperforms the other methods. Furthermore, BCL achieves better performance on long-tailed CIFAR datasets with large imbalance factors. We note that the accuracy gap between BCL and Hybrid-SC decreases as the degree of data imbalance reduces. This result is mainly attributed to the fact that the conventional supervised contrastive loss leads to more serious bias in representation learning when unbalance problem is more severe. 

Further, we report the accuracy on three groups of classes, including Many-shot($>$100 images), Medium-shot(20 $\sim$ 100 images), Few-shot($<$20 images), on CIFAR-100-LT with imbalance factor as 100. Additionally, for a fair comparison with~\cite{paco}, we report the accuracy on 200 and 400 epochs. 
DRO-LT~\cite{dro-lt} is also a contrastive learning method which extends prototypical contrastive learning by introducing distributional robustness. As shown in Table~\ref{tab:cifar100-100}, BCL surpasses DRO-LT by 4.6 and PaCo by 1.9 with 200 training epochs and 400 training epochs, respectively. It's worthwhile to mention that different to most of the previous methods which compromise the performance of the head classes, our BCL further improves the performance of the head classes while simultaneously improving the tails. 
% Compared with the ensemble-based model RIDE,  BCL consistently demonstrates better recognition performance.

\noindent \textbf{ImageNet-LT}\quad
Table~\ref{tab:inat-img} and Table~\ref{tab:imagenet-lt-top1-resnext} list the results on ImageNet-LT. We report the overall Top-1 accuracy as well as the Top-1 accuracy on Many-shot, Medium-shot, and Few-shot groups. Compared with Balanced Softmax~\cite{balancedsoftmax}, which proposes a logit compensation by adjusting the predictions according to the class frequency. BCL significantly outperforms Balanced Softmax on all groups, confirming the well-learned representations can boost the overall performance. LWS~\cite{decoupling}, $\tau$-norm~\cite{decoupling}, and DisAlign~\cite{distributionalign} adopt the two-stage learning strategy. These methods focus on fine-tuning the classifier in the second stage, while they neglect the bias implied in the representation learning stage. PaCo~\cite{paco} uses a set of parametric centers in supervised contrastive learning. These centers are assigned with a much greater weight, which can be regarded as the weights of a classifier. However, the prototypes used in BCL complement the samples of each class to make sure that all classes appear in every mini-batch. Compared with PaCo, BCL achieves a better overall accuracy of 57.1$\%$ with remarkable accuracy improvements on the head and few classes.
\begin{table}[t]
    \centering
    \small
   \begin{tabular}{l|| c | c }
    \hline
    
         Methods & ImageNet-LT & iNaturalist 2018 \\
         \hline
        $\tau$-norm\cite{decoupling} & 46.7 & 65.6\\
        cRT\cite{decoupling} & 49.6 & 65.2\\
        LWS\cite{decoupling} & 49.9 & 65.9\\
        %  Causual Model\cite{casual} & 51.8 & 63.9\\
        BBN\cite{bbn} & - & 66.3\\
        Hybrid-SC\cite{hybrid} & - & 66.7\\
        Hybrid-PSC\cite{hybrid} & - & 68.1\\
        SSP\cite{rethinking} & 51.3 & 68.1\\
        KCL\cite{kcl} & 51.5 & 68.6\\
        DisAlign\cite{distributionalign} & 52.9 & 69.5\\
        RIDE(2 experts)\cite{ride} & 54.4 & 71.4\\
        \hline
        BCL(Ours) & \textbf{56.0} & \textbf{71.8}\\
         \hline
\end{tabular}
    \caption{Top-1 accuracy of ResNet-50 on ImageNet-LT and iNaturalist 2018. All methods are trained for up to 100 epochs. }
    \label{tab:inat-img}
\end{table}

\noindent \textbf{iNaturalist 2018}\quad
Table~\ref{tab:inat-img} shows the experimental results on iNaturalist 2018. Since BCL is a contrastive learning method, it benefits more from a longer training time. However, for a fair comparison, we report the results of various models trained for up to 100 epochs. Hybrid-SC~\cite{hybrid} and Hybrid-PSC~\cite{hybrid} are contrastive learning approaches, and their performances are inferior to BCL due to the potential bias produced in representation learning. RIDE~\cite{ride} trains model from an ensemble-based manner.
%and achieves quite excellent performance on several long-tailed benchmarks. 
RIDE obtains high performance by routing diverse experts. However, it's different from our motivation of boosting long-tailed learning via training a sufficient encoder and a balanced feature space. Compared with the ensemble-based model RIDE, BCL consistently demonstrates better recognition performance and achieves the overall accuracy at 71.8$\%$.

% \noindent \textbf{Comparison to state-of-the-art methods}\quad
% In this part, we discuss our proposed BCL compared to the existing state-of-the-art methods. 
% RIDE~\cite{ride} trains model from an ensemble-based manner and achieves quite excellent performance on several long-tailed benchmarks. RIDE obtains high performance by routing diverse experts and distillation, however, this is different from our motivation of boosting long-tailed learning via training a sufficient encoder and a balanced feature space. As shown in these experiments, BCL also achieve competitive results compared to ensemble-based methods.
% PaCo~\cite{paco} uses parametric centers for contrastive learning which can be regarded as a classifier. PaCo distinguishes between the center and samples by weighting, while the prototypes in BCL supplements the samples of a class so that all classes within the batch appear.
% In addition, BCL ensures that the loss of each sample is consistent and class-independent when the lower bound is attained, which implies that the learning will not be biased towards the head classes.
\begin{table}[t]
    \centering
    \small
   \begin{tabular}{l||cccc}
    \hline
    Methods & Many & Medium & Few & All \\
         \hline
        90 epochs\\
        Focal Loss$^\dagger$\cite{focal} & 64.3 & 37.1 & 8.2 &43.7\\
        $\tau$-norm\cite{decoupling} & 59.1 & 46.9 & 30.7 &49.4\\
        Balanced Softmax$^\dagger$\cite{balancedsoftmax} & 62.2 & 48.8 & 29.8 & 51.4\\
        LWS\cite{decoupling} & 60.2 & 47.2 & 30.3 & 49.9\\
        LADE\cite{disentangling} & 62.3 & 49.3 & 31.2 & 51.9 \\
        Casual model\cite{casual} & 62.7 & 48.8 & 31.6 & 51.8\\
        DisAlign\cite{distributionalign}  & 62.7 & 52.1 & 31.4& 53.4\\
        RIDE(2 experts)\cite{ride} & - & - & - & 55.9 \\
        \hline
        BCL(Ours) & \textbf{67.2} & \textbf{53.9} & \textbf{36.5} & \textbf{56.7}\\
        \hline
        180 epochs\\
        LADE\cite{disentangling} & 65.1 & 48.9 & 33.4 & 53.0\\
        Balanced Softmax$^\dagger$ & 65.8 & 53.2 & 34.1 & 55.4\\
        
%        RIDE & 68.0 & 52.9 & 35.1 & 56.3\\
%        TADE & 66.5 & 57.0 & 43.5 & 58.8\\
        PaCo$^\dagger$\cite{paco} & 64.4 & \textbf{55.7} & 33.7 & 56.0\\
        \hline
        BCL(Ours) & \textbf{67.9} & 54.2 & \textbf{36.6} & \textbf{57.1} \\
        \hline
\end{tabular}
    %\caption{Top-1 accuracy of ResNext-50 on ImageNet-LT. For a fair comparison, we report results of 90 epochs and 180 epochs. $\dagger$ denotes results borrowed from \cite{disentangling} and
    %* denotes results reproduced for 180 epochs using code published by authors.}
    \caption{Top-1 accuracy of ResNext-50 on ImageNet-LT. We report the results of 90 epochs and 180 epochs. $\dagger$ denotes results reproduced using the code released by authors, both with RandAug.}
    \label{tab:imagenet-lt-top1-resnext}
\end{table}

% \begin{table}[h]
%     \centering
%   \begin{tabular}{l c c c c}
%     \hline
%          Methods & Many & Medium & Few & All  \\
%          \hline
%          CE & 45.7 & 27.3 & 8.2 & 30.2\\ 
%         $\tau$ -norm & 37.8 & 40.7 & 31.8 & 37.9\\
%         Balanced Softmax & 42.0 & 39.3 & 30.5 & 38.6\\
%         LADE & 42.8&39.0&31.2&38.8\\
%         DisAlign & 40.4&42.4&30.1&39.3\\
%         MiSLAS & 39.6& 43.3 & 36.1 & 40.4\\
%         PaCo & 37.5 & 47.2 & 33.9&41.2\\
%         \hline
%         BCL(Ours) & -\\
%          \hline
% \end{tabular}
%     \caption{Top-1 accuracy over all classes of ResNet-50 on iNaturalist 2018. By default, the methods are trained up for 100 epochs. }
%     \label{tab:places}
% \end{table}
\section{Conclusion}
% In this work, we have presented a framework unifying a classification branch and a contrastive learning branch to address the long-tailed visual recognition problem. To tackle the imbalanced representation space and improve the representation ability, we develop a balanced contrastive learning method to make the lower bound of loss independent from the class information. Ultimately, all classes are optimized for a regular simplex configuration that yields a balanced feature space. 
% In addition to the balanced contrastive learning branch, we employ a classification branch with logit compensation to tackle the biased classifier. We conduct extensive experiments on the long-tailed benchmarks of long-tailed CIFAR, ImageNet-LT, and iNaturalist 2018. The experimental results adequately demonstrate the superiority of BCL compared to existing long-tailed learning methods.
In this work, we investigated the problem of long-tailed recognition from the perspective of representation learning. We provided in-depth analysis to demonstrate that existing supervised contrastive learning forms an undesired asymmetric geometry configuration for long-tailed data. To tackle the imbalanced data representation learning problem, we developed a balanced contrastive loss, so that
%, whose lower bound is independent of the class distribution. With the balanced contrastive loss, 
all classes are optimized for a regular simplex configuration that yields a balanced feature space.
%To tackle the imbalanced representation space, we presented balanced contrastive learning. Specifically, we develop a balanced contrastive loss, whose lower bound is independent of the class information. Ultimately, all classes are optimized for a regular simplex configuration that yields a balanced feature space.
%The proposed balanced contrastive learning is easy to combine with existing classifier debias works, 
In addition to BCL, we employ a classification branch with logit compensation to tackle the biased classifier. Overall we have presented a framework unifying both branches. We conducted extensive experiments on the long-tailed benchmarks of long-tailed CIFAR, ImageNet-LT, and iNaturalist 2018. The experimental results adequately demonstrate the superiority of BCL compared to existing long-tailed learning methods.

\section{Acknowledgements}
This work was supported in part by NSFC project (\# 62072116), Shanghai Municipal Commission of Economy and Informatization Project (2020-GYHLW-01009), and in part by Shanghai Pujiang Program (20PJ1401900).

%%%%%%%%% REFERENCES
{\small
\bibliographystyle{ieee_fullname}
\bibliography{egbib}

\begin{thebibliography}{10}\itemsep=-1pt

\bibitem{oversample_systematic}
Mateusz Buda, Atsuto Maki, and Maciej~A Mazurowski.
\newblock A systematic study of the class imbalance problem in convolutional
  neural networks.
\newblock {\em Neural Networks}, 106:249--259, 2018.

\bibitem{oversample_effect}
Jonathon Byrd and Zachary Lipton.
\newblock What is the effect of importance weighting in deep learning?
\newblock In {\em International Conference on Machine Learning}, pages
  872--881. PMLR, 2019.

\bibitem{LDAM}
Kaidi Cao, Colin Wei, Adrien Gaidon, Nikos Arechiga, and Tengyu Ma.
\newblock Learning imbalanced datasets with label-distribution-aware margin
  loss.
\newblock In {\em Proceedings of the 33rd International Conference on Neural
  Information Processing Systems}, pages 1567--1578, 2019.

\bibitem{chen2020study}
Jingjing Chen, Bin Zhu, Chong-Wah Ngo, Tat-Seng Chua, and Yu-Gang Jiang.
\newblock A study of multi-task and region-wise deep learning for food
  ingredient recognition.
\newblock {\em IEEE Transactions on Image Processing}, 30:1514--1526, 2020.

\bibitem{simclr}
Ting Chen, Simon Kornblith, Mohammad Norouzi, and Geoffrey Hinton.
\newblock A simple framework for contrastive learning of visual
  representations.
\newblock In {\em International conference on machine learning}, pages
  1597--1607. PMLR, 2020.

\bibitem{autoaugment}
Ekin~D Cubuk, Barret Zoph, Dandelion Mane, Vijay Vasudevan, and Quoc~V Le.
\newblock Autoaugment: Learning augmentation strategies from data.
\newblock In {\em Proceedings of the IEEE/CVF Conference on Computer Vision and
  Pattern Recognition}, pages 113--123, 2019.

\bibitem{reslt}
Jiequan Cui, Shu Liu, Zhuotao Tian, Zhisheng Zhong, and Jiaya Jia.
\newblock Reslt: Residual learning for long-tailed recognition.
\newblock {\em arXiv preprint arXiv:2101.10633}, 2021.

\bibitem{paco}
Jiequan Cui, Zhisheng Zhong, Shu Liu, Bei Yu, and Jiaya Jia.
\newblock Parametric contrastive learning.
\newblock In {\em Proceedings of the IEEE/CVF International Conference on
  Computer Vision}, pages 715--724, 2021.

\bibitem{CBLoss}
Yin Cui, Menglin Jia, Tsung-Yi Lin, Yang Song, and Serge Belongie.
\newblock Class-balanced loss based on effective number of samples.
\newblock In {\em Proceedings of the IEEE/CVF conference on computer vision and
  pattern recognition}, pages 9268--9277, 2019.

\bibitem{cbfocal}
Yin Cui, Menglin Jia, Tsung-Yi Lin, Yang Song, and Serge Belongie.
\newblock Class-balanced loss based on effective number of samples.
\newblock In {\em Proceedings of the IEEE/CVF conference on computer vision and
  pattern recognition}, pages 9268--9277, 2019.

\bibitem{imagenet}
Jia Deng, Wei Dong, Richard Socher, Li-Jia Li, Kai Li, and Li Fei-Fei.
\newblock Imagenet: A large-scale hierarchical image database.
\newblock In {\em 2009 IEEE conference on computer vision and pattern
  recognition}, pages 248--255. Ieee, 2009.

\bibitem{deng2019mixed}
Lixi Deng, Jingjing Chen, Qianru Sun, Xiangnan He, Sheng Tang, Zhaoyan Ming,
  Yongdong Zhang, and Tat~Seng Chua.
\newblock Mixed-dish recognition with contextual relation networks.
\newblock In {\em Proceedings of the 27th ACM International Conference on
  Multimedia}, pages 112--120, 2019.

\bibitem{cutout}
Terrance DeVries and Graham~W Taylor.
\newblock Improved regularization of convolutional neural networks with cutout.
\newblock {\em arXiv preprint arXiv:1708.04552}, 2017.

\bibitem{undersample}
Chris Drumnond and Robert~C Holte.
\newblock Class imbalance and cost sensitivity: Why undersampling beats
  oversampling.
\newblock In {\em ICML-KDD 2003 Workshop: Learning from Imbalanced Datasets},
  volume~3, 2003.

\bibitem{fang2021exploring}
Cong Fang, Hangfeng He, Qi Long, and Weijie~J Su.
\newblock Exploring deep neural networks via layer-peeled model: Minority
  collapse in imbalanced training.
\newblock {\em Proceedings of the National Academy of Sciences}, 118(43), 2021.

\bibitem{dissecting}
Florian Graf, Christoph Hofer, Marc Niethammer, and Roland Kwitt.
\newblock Dissecting supervised constrastive learning.
\newblock In {\em International Conference on Machine Learning}, pages
  3821--3830. PMLR, 2021.

\bibitem{han2020self}
Tengda Han, Weidi Xie, and Andrew Zisserman.
\newblock Self-supervised co-training for video representation learning.
\newblock {\em Advances in Neural Information Processing Systems},
  33:5679--5690, 2020.

\bibitem{moco}
Kaiming He, Haoqi Fan, Yuxin Wu, Saining Xie, and Ross Girshick.
\newblock Momentum contrast for unsupervised visual representation learning.
\newblock In {\em Proceedings of the IEEE/CVF Conference on Computer Vision and
  Pattern Recognition}, pages 9729--9738, 2020.

\bibitem{resnet}
Kaiming He, Xiangyu Zhang, Shaoqing Ren, and Jian Sun.
\newblock Deep residual learning for image recognition.
\newblock In {\em Proceedings of the IEEE conference on computer vision and
  pattern recognition}, pages 770--778, 2016.

\bibitem{disentangling}
Youngkyu Hong, Seungju Han, Kwanghee Choi, Seokjun Seo, Beomsu Kim, and Buru
  Chang.
\newblock Disentangling label distribution for long-tailed visual recognition.
\newblock In {\em Proceedings of the IEEE/CVF Conference on Computer Vision and
  Pattern Recognition}, pages 6626--6636, 2021.

\bibitem{reweight_learning}
Chen Huang, Yining Li, Chen~Change Loy, and Xiaoou Tang.
\newblock Learning deep representation for imbalanced classification.
\newblock In {\em Proceedings of the IEEE conference on computer vision and
  pattern recognition}, pages 5375--5384, 2016.

\bibitem{kcl}
Bingyi Kang, Yu Li, Sa Xie, Zehuan Yuan, and Jiashi Feng.
\newblock Exploring balanced feature spaces for representation learning.
\newblock In {\em International Conference on Learning Representations}, 2020.

\bibitem{decoupling}
Bingyi Kang, Saining Xie, Marcus Rohrbach, Zhicheng Yan, Albert Gordo, Jiashi
  Feng, and Yannis Kalantidis.
\newblock Decoupling representation and classifier for long-tailed recognition.
\newblock In {\em International Conference on Learning Representations}, 2019.

\bibitem{scl}
Prannay Khosla, Piotr Teterwak, Chen Wang, Aaron Sarna, Yonglong Tian, Phillip
  Isola, Aaron Maschinot, Ce Liu, and Dilip Krishnan.
\newblock Supervised contrastive learning.
\newblock {\em Advances in Neural Information Processing Systems}, 33, 2020.

\bibitem{alexnet}
Alex Krizhevsky, Ilya Sutskever, and Geoffrey~E Hinton.
\newblock Imagenet classification with deep convolutional neural networks.
\newblock {\em Advances in neural information processing systems},
  25:1097--1105, 2012.

\bibitem{PCL}
Junnan Li, Pan Zhou, Caiming Xiong, and Steven~C.H. Hoi.
\newblock Prototypical contrastive learning of unsupervised representations.
\newblock In {\em ICLR}, 2021.

\bibitem{metasaug}
Shuang Li, Kaixiong Gong, Chi~Harold Liu, Yulin Wang, Feng Qiao, and Xinjing
  Cheng.
\newblock Metasaug: Meta semantic augmentation for long-tailed visual
  recognition.
\newblock In {\em Proceedings of the IEEE/CVF Conference on Computer Vision and
  Pattern Recognition}, pages 5212--5221, 2021.

\bibitem{li2021targeted}
Tianhong Li, Peng Cao, Yuan Yuan, Lijie Fan, Yuzhe Yang, Rogerio Feris, Piotr
  Indyk, and Dina Katabi.
\newblock Targeted supervised contrastive learning for long-tailed recognition.
\newblock {\em arXiv preprint arXiv:2111.13998}, 2021.

\bibitem{focal}
Tsung-Yi Lin, Priya Goyal, Ross Girshick, Kaiming He, and Piotr Doll{\'a}r.
\newblock Focal loss for dense object detection.
\newblock In {\em Proceedings of the IEEE international conference on computer
  vision}, pages 2980--2988, 2017.

\bibitem{largescale}
Ziwei Liu, Zhongqi Miao, Xiaohang Zhan, Jiayun Wang, Boqing Gong, and Stella~X
  Yu.
\newblock Large-scale long-tailed recognition in an open world.
\newblock In {\em Proceedings of the IEEE/CVF Conference on Computer Vision and
  Pattern Recognition}, pages 2537--2546, 2019.

\bibitem{logitadjustment}
Aditya~Krishna Menon, Sadeep Jayasumana, Ankit~Singh Rawat, Himanshu Jain,
  Andreas Veit, and Sanjiv Kumar.
\newblock Long-tail learning via logit adjustment.
\newblock In {\em International Conference on Learning Representations}, 2020.

\bibitem{prevalence}
Vardan Papyan, XY Han, and David~L Donoho.
\newblock Prevalence of neural collapse during the terminal phase of deep
  learning training.
\newblock {\em Proceedings of the National Academy of Sciences},
  117(40):24652--24663, 2020.

\bibitem{oversample_dynamic}
Samira Pouyanfar, Yudong Tao, Anup Mohan, Haiman Tian, Ahmed~S Kaseb, Kent
  Gauen, Ryan Dailey, Sarah Aghajanzadeh, Yung-Hsiang Lu, Shu-Ching Chen,
  et~al.
\newblock Dynamic sampling in convolutional neural networks for imbalanced data
  classification.
\newblock In {\em 2018 IEEE conference on multimedia information processing and
  retrieval (MIPR)}, pages 112--117. IEEE, 2018.

\bibitem{balancedsoftmax}
Jiawei Ren, Cunjun Yu, Shunan Sheng, Xiao Ma, Haiyu Zhao, Shuai Yi, and
  Hongsheng Li.
\newblock Balanced meta-softmax for long-tailed visual recognition.
\newblock In {\em Proceedings of Neural Information Processing
  Systems(NeurIPS)}, Dec 2020.

\bibitem{fasterrcnn}
Shaoqing Ren, Kaiming He, Ross Girshick, and Jian Sun.
\newblock Faster r-cnn: Towards real-time object detection with region proposal
  networks.
\newblock {\em Advances in neural information processing systems}, 28:91--99,
  2015.

\bibitem{dro-lt}
Dvir Samuel and Gal Chechik.
\newblock Distributional robustness loss for long-tail learning.
\newblock In {\em Proceedings of the IEEE/CVF International Conference on
  Computer Vision (ICCV)}, pages 9495--9504, October 2021.

\bibitem{su2020video}
Zixuan Su, Xindi Shang, Jingjing Chen, Yu-Gang Jiang, Zhiyong Qiu, and Tat-Seng
  Chua.
\newblock Video relation detection via multiple hypothesis association.
\newblock In {\em Proceedings of the 28th ACM International Conference on
  Multimedia}, pages 3127--3135, 2020.

\bibitem{equalization}
Jingru Tan, Changbao Wang, Buyu Li, Quanquan Li, Wanli Ouyang, Changqing Yin,
  and Junjie Yan.
\newblock Equalization loss for long-tailed object recognition.
\newblock In {\em Proceedings of the IEEE/CVF conference on computer vision and
  pattern recognition}, pages 11662--11671, 2020.

\bibitem{casual}
Kaihua Tang, Jianqiang Huang, and Hanwang Zhang.
\newblock Long-tailed classification by keeping the good and removing the bad
  momentum causal effect.
\newblock {\em Advances in Neural Information Processing Systems}, 33, 2020.

\bibitem{inaturalist}
Grant Van~Horn, Oisin Mac~Aodha, Yang Song, Yin Cui, Chen Sun, Alex Shepard,
  Hartwig Adam, Pietro Perona, and Serge Belongie.
\newblock The inaturalist species classification and detection dataset.
\newblock In {\em Proceedings of the IEEE conference on computer vision and
  pattern recognition}, pages 8769--8778, 2018.

\bibitem{understandingssl}
Feng Wang and Huaping Liu.
\newblock Understanding the behaviour of contrastive loss.
\newblock In {\em Proceedings of the IEEE/CVF Conference on Computer Vision and
  Pattern Recognition}, pages 2495--2504, 2021.

\bibitem{hybrid}
Peng Wang, Kai Han, Xiu-Shen Wei, Lei Zhang, and Lei Wang.
\newblock Contrastive learning based hybrid networks for long-tailed image
  classification.
\newblock In {\em Proceedings of the IEEE/CVF Conference on Computer Vision and
  Pattern Recognition}, pages 943--952, 2021.

\bibitem{wang2022cross}
Rui Wang, Zuxuan Wu, Zejia Weng, Jingjing Chen, Guo-Jun Qi, and Yu-Gang Jiang.
\newblock Cross-domain contrastive learning for unsupervised domain adaptation.
\newblock {\em IEEE Transactions on Multimedia}, 2022.

\bibitem{ride}
Xudong Wang, Long Lian, Zhongqi Miao, Ziwei Liu, and Stella Yu.
\newblock Long-tailed recognition by routing diverse distribution-aware
  experts.
\newblock In {\em International Conference on Learning Representations}, 2020.

\bibitem{reweight_learningtotail}
Yu-Xiong Wang, Deva Ramanan, and Martial Hebert.
\newblock Learning to model the tail.
\newblock In {\em Proceedings of the 31st International Conference on Neural
  Information Processing Systems}, pages 7032--7042, 2017.

\bibitem{wang2021visual}
Zheng Wang, Jingjing Chen, and Yu-Gang Jiang.
\newblock Visual co-occurrence alignment learning for weakly-supervised video
  moment retrieval.
\newblock In {\em Proceedings of the 29th ACM International Conference on
  Multimedia}, pages 1459--1468, 2021.

\bibitem{resnext}
Saining Xie, Ross Girshick, Piotr Doll{\'a}r, Zhuowen Tu, and Kaiming He.
\newblock Aggregated residual transformations for deep neural networks.
\newblock In {\em Proceedings of the IEEE conference on computer vision and
  pattern recognition}, pages 1492--1500, 2017.

\bibitem{rethinking}
Yuzhe Yang and Zhi Xu.
\newblock Rethinking the value of labels for improving class-imbalanced
  learning.
\newblock In {\em Conference on Neural Information Processing Systems
  (NeurIPS)}, 2020.

\bibitem{zhang2021token}
Hao Zhang, Yanbin Hao, and Chong-Wah Ngo.
\newblock Token shift transformer for video classification.
\newblock In {\em Proceedings of the 29th ACM International Conference on
  Multimedia}, pages 917--925, 2021.

\bibitem{distributionalign}
Songyang Zhang, Zeming Li, Shipeng Yan, Xuming He, and Jian Sun.
\newblock Distribution alignment: A unified framework for long-tail visual
  recognition.
\newblock In {\em Proceedings of the IEEE/CVF Conference on Computer Vision and
  Pattern Recognition}, pages 2361--2370, 2021.

\bibitem{bbn}
Boyan Zhou, Quan Cui, Xiu-Shen Wei, and Zhao-Min Chen.
\newblock Bbn: Bilateral-branch network with cumulative learning for
  long-tailed visual recognition.
\newblock In {\em Proceedings of the IEEE/CVF Conference on Computer Vision and
  Pattern Recognition}, pages 9719--9728, 2020.

\end{thebibliography}
}

\pagebreak
%\begin{widetext}
\onecolumn
\setcounter{equation}{0}
\setcounter{figure}{0}
\setcounter{table}{0}
\setcounter{page}{1}
\setcounter{section}{0}
\makeatletter

\begin{center}
\textbf{\huge Supplementary Material}
\end{center}
\section{Proof of Theorem 2 and Theorem 3}
In this section, we will proof Theorem 2 and Theorem 3 proposed in section 3.2. The main idea of the proof is to decouple the class-specific batch-wise loss as attraction term and repulsion term as in~\cite{dissecting}. First, we will show the spontaneous appearance of variability collapse as the training process in attraction term. When this condition holds, we find that to minimize the loss, the solution of the model spontaneously satisfies the simplex configuration. 

Before the detailed derivation, recall the main notions and definitions of this paper:
\begin{itemize}
    \item $h,K,N \in \mathbb{N}$
    \item $\mathcal{Z}=\mathbb{R}^{h}$
    \item $\mathcal{Y}=[K]=\{1,2,…,K\}$
\end{itemize}
\noindent \textbf{Definition 1 (Supervise contrastive loss)}\quad
For an instance $x_i$ of representation $z_i$ in a batch $B$, supervised contrastive loss has the following expression:
\begin{equation}
    \mathcal{L}_{i}=-\frac{1}{\vert B_{y}\vert-1}\sum_{p\in B_{y}\setminus\lbrace{i}\rbrace}\log\frac{\exp(z_{i}\cdot z_{p})}{\sum\limits_{k\in B\setminus\{i\}}\exp(z_{i}\cdot z_{k})}
\end{equation}
\noindent \textbf{Definition 2 (Balanced contrastive loss)}\quad
For an instance $x_i$ of representation $z_i$ in a batch $B$, balanced contrastive loss has the following expression:
\begin{equation}
    \mathcal{L}_{i}=-\frac{1}{\vert B_{y}\vert -1}\sum_{p\in B_{y}\setminus\lbrace{i}\rbrace}\log\frac{\exp(z_{i}\cdot z_{p})}{\sum \limits_{j\in \mathcal{Y}_{B}}\frac{1}{\vert B_{j}\vert}\sum \limits_{k\in B_{j}}\exp(z_{i}\cdot z_{k})}
\end{equation}

\noindent \textbf{Definition 3 (Class-specific batch-wise loss)}
\begin{equation}
    \mathcal{L}(Z;Y,B,y) = \begin{cases}
    \sum_{i\in B_{y}} \mathcal{L}_{i} & if \  \vert B_{y}\vert > 1\\
    0 & else\\
    \end{cases}
\end{equation}
\noindent \textbf{Definition 4 (Regular simplex)}\quad
A set of points $\zeta_{1},\dots,\zeta_{K}\in \mathbb{R}^{h}$ form the vertices of a regular simplex inscribed in the hypersphere of radius $\rho>0$, if and only if the following conditions hold:
\begin{enumerate}[(1)]
    \item $\sum\nolimits_{i\in[K]}\zeta_{i}=0$
    \item $\Vert \zeta_{i} \Vert=\rho, \text{for} \ i \in {[K]}$
    \item $\exists d \in \mathbb{R}: d = \langle\zeta_{i},\zeta_{j}\rangle\ \text{for}\ 1\leq i < j \leq K$
\end{enumerate}
where $B_{y}$ and $\mathcal{Y}_B$ are  subsets of $B$ and $\mathcal{Y}$, respectively. Note that the $\vert B_{j} \vert$ term in above equations will minus one when the positive class is averaged. Here, we omit hyper parameter temperature $\tau$ and $\langle \cdot \rangle$ for the inner product operation.
Additionally, we default to $K\geq h+1$ and assume $\Vert z_{i} \Vert_{2}=1$.

\noindent \textbf{Proof of Theorem 2}\quad
First we rewrite class-specific batch-wise loss as following form:
\begin{equation} \label{eq:3}
\begin{aligned}
    \mathcal{L}_{BCL}(Z;Y,B,y) &= 
    \sum_{i\in B_{y}} -\frac{1}{\vert B_{y}\vert -1}\sum_{p\in B_{y}\setminus\lbrace{i}\rbrace}\log\frac{\exp(z_{i}\cdot z_{p})}{\sum\limits_{j\in \mathcal{Y}_{B}}\frac{1}{\vert B_{j}\vert}\sum\limits_{k\in B_{j}}\exp(z_{i}\cdot z_{k})} \\
    &=\sum_{i\in B_{y}}\log \left( \frac{
    \sum\limits_{j\in \mathcal{Y}_{B}}\frac{1}{\vert B_{j}\vert}\sum\limits_{k\in B_{j}}\exp(z_{i}\cdot z_{k})
    }{\prod\limits_{p\in B_{y}\setminus\lbrace{i}\rbrace}
     \exp ( z_{i}, z_{p} )^{1 / \vert B_{y}\vert -1}}
    \right) \\
    &=\sum_{i\in B_{y}}\log \left( \frac{
    \sum\limits_{j\in \mathcal{Y}_{B}}\frac{1}{\vert B_{j}\vert}\sum\limits_{k\in B_{j}}\exp(z_{i}\cdot z_{k})
    }{\exp ( \frac{1}{\vert B_{y}\vert -1} \sum\limits_{p\in B_{y}\setminus\lbrace{i}\rbrace} z_{i}\cdot z_{p} )}
    \right)
\end{aligned}
\end{equation}
The key idea is to divide the sum in the numerator into positives and negatives. Since the exponential function is convex, by applying Jensen's inequality, we have
\begin{equation}
\begin{aligned}
    \frac{1}{\vert B_{y}\vert -1}
    \sum\limits_{k\in B_{y}\setminus\lbrace{i}\rbrace}\exp(z_{i}\cdot z_{k}) &\overset{(Q1)}{\geq} \exp \left( \frac{1}{\vert B_{y}\vert -1}
    \sum\limits_{k\in B_{y}\setminus\lbrace{i}\rbrace} z_{i} \cdot z_{k}
    \right) \\
      \frac{1}{\vert B_{j}\vert }
    \sum\limits_{k\in B_{j} \atop j \neq y}\exp(z_{i}\cdot z_{k}) &\overset{(Q2)}{\geq} \exp \left( \frac{1}{\vert B_{j}\vert}
    \sum\limits_{k\in B_{j}} z_{i} \cdot z_{k}
    \right)
\end{aligned}
\end{equation}
The equality is attained if and only if:
\begin{enumerate}[(Q1)]
    \item There is $C_{i}(B,y)$ such that $\forall k \in B_{y}\setminus\lbrace {i}\rbrace$ all inner products $z_{i} \cdot z_{k}=C_{i}(B,y)$ are equal.
    \item There is $D_{i}(B,y,j)$ such that $\forall k \in B_{j,j\neq y}$ all inner products $z_{i} \cdot z_{k}=D_{i}(B,y,j)$ are equal.
\end{enumerate}
Thus, the sum in the numerator can be written as follows
\begin{equation}
     \sum\limits_{j\in \mathcal{Y}_{B}}\frac{1}{\vert B_{j}\vert}\sum\limits_{j\in B_{j}}\exp(z_{i}\cdot z_{k}) \geq \exp \left( \frac{1}{\vert B_{y}\vert -1}
    \sum\limits_{p\in B_{y}\setminus\lbrace{i}\rbrace} z_{i} \cdot z_{p}
    \right) + \sum\limits_{j \in \mathcal{Y}_{B} \atop j \neq y}
    \exp \left( \frac{1}{\vert B_{j}\vert}
    \sum\limits_{k\in B_{j}} z_{i} \cdot z_{k} \right)
\end{equation}
By leverage Jensen's inequality again on the latter term, resulting in
\begin{equation}
    \sum\limits_{j \in \mathcal{Y}_{B} \atop j \neq y}
    \exp \left( \frac{1}{\vert B_{j}\vert}
    \sum\limits_{k\in B_{j}} z_{i} \cdot z_{k} \right)
    \overset{(Q3)}{\geq} ( \vert \mathcal{Y}_{B} \vert -1 ) \exp \left(
    \frac{1}{\vert \mathcal{Y}_{B} \vert -1 } 
     \sum\limits_{j \in \mathcal{Y}_{B} \atop j \neq y}
    \frac{1}{\vert B_{j}\vert}
    \sum\limits_{k\in B_{j}} z_{i} \cdot z_{k}
    \right)
\end{equation}
Here, the equality is attained if and only if
\begin{enumerate}[(Q3)]
    \item There is $E_{i}(B,y)$ such that $\forall j \in \mathcal{Y}_{B,j\neq y}, \forall k \in B_{j}$ all inner products $z_{i} \cdot z_{k}=E_{i}(B,y)$ are equal.
\end{enumerate}
Thus, for a specific mini-batch, Eq.~\ref{eq:3} can be written as 
\begin{equation}
\begin{aligned}
    \mathcal{L}_{BCL}(Z;Y,B,y) \geq \sum_{i\in B_{y}}\log \left(
    1+(\vert \mathcal{Y}_{B}\vert -1) \exp \left(
    \underbrace{
    \frac{1}{\vert \mathcal{Y}_{B}\vert -1} \sum_{q\in\mathcal{Y}_{B}\setminus{\lbrace y \rbrace}} \frac{1}{\vert B_{q}\vert}
    \sum_{k \in B_{q} }
    z_{i}\cdot z_{k}}_{repulsion\ term}
    \underbrace{-
     \frac{1}{\vert B_{y}\vert -1} \sum_{j\in B_{y}\setminus{\lbrace i \rbrace}} z_{i}\cdot z_{j}}_{attraction\ term}\right)\right)
\end{aligned}
\end{equation}
which ends the proof of Theorem 2. Here, the equality is attained if and only if conditions (Q1) and (Q3) hold for every $i\in B_{y}$. Additionally, constants $C_{i}(B,y)$ and $E_{i}(B,y)$ only depend on the batch $B$ and the label $y$.

\noindent \textbf{Proof of Theorem 3}\quad
On the basis of theorem 2, we assume $\mathcal{Y}_{B}=\mathcal{Y}$ for every batch $B$.
For simplicity, we rewrite the two terms of the exponential function in Theorem 2 as the following form
\begin{equation}
\begin{aligned}
    S(Z;Y,B,y)&=S_{att}(Z;Y,B,y) + S_{rep}(Z;Y,B,y)\\
    S_{att}(Z;Y,B,y) &= - \frac{1}{\vert B_{y}\vert -1} \sum_{j\in B_{y}\setminus{\lbrace i \rbrace}} z_{i}\cdot z_{j} \\
    S_{rep}(Z;Y,B,y) &= \frac{1}{\vert \mathcal{Y}\vert -1} \sum_{q\in\mathcal{Y}\setminus{\lbrace y \rbrace}} \frac{1}{\vert B_{q}\vert}
    \sum_{k \in B_{q} }
    z_{i}\cdot z_{k}
\end{aligned}
\end{equation}
Regroup the addends, we can obtain the following formulation
\begin{equation}
\begin{aligned}
    \mathcal{L}_{BCL}(Z;Y)&=\sum_{B \in \mathcal{B}} \sum_{y \in \mathcal{Y}}
    \mathcal{L}_{BCL}(Z;Y,B,y) \\
    &\geq \sum_{B \in \mathcal{B}} \sum_{y \in \mathcal{Y}} \sum_{i \in B_{y}} \log \left( 1+ (\vert \mathcal{Y} \vert -1) \exp 
    ( S(Z;Y,B,y) )
    \right)
\end{aligned}
\end{equation}
Let $\alpha > 0$, and $f:\mathbb{R}\rightarrow \mathbb{R}, x \rightarrow \log(1+\alpha \exp(x))$. It is easy to verify that the function $f$ is smooth with second derivative and convex. According to Jensen's inequality, we obtain the lower bound as follows
\begin{equation}
    \mathcal{L}_{BCL}(Z;Y) \overset{(Q4)}{\geq} \vert \mathcal{D} \vert \log \left(1+
    (\vert \mathcal{Y} \vert -1 ) \exp \left(
    \sum_{B \in \mathcal{B}} \sum_{y \in \mathcal{Y}} \sum_{i \in {B}_{y}} S(Z;Y,B,y) \right) \right)
\end{equation}
where $\mathcal{D}$ denotes the dataset, the equality is attained if and only if:
\begin{enumerate}[(Q4)]
    \item There is constant $\theta$ such that $\forall B \in \mathcal{B}, \forall y \in \mathcal{Y}$ and $\forall i \in B_{y}$, the values of $S(Z;Y,B,y)=\theta$ agree.
\end{enumerate}
Next we derive the sum of attraction terms. For every $Y \in \mathcal{Y}^{N}$ and every $Z \in \mathcal{Z}^{N}$, using the Cauchy-Schwarz inequality and the assumption that $\mathcal{Z}$ is a unit hypersphere, we have
\begin{equation}
\begin{aligned}
    \sum_{i \in {B}_{y}} S_{att}(Z;Y,B,y) &= - \frac{1}{\vert B_{y}\vert -1}
    \sum_{i \in {B}_{y}}
    \sum_{j\in B_{y}\setminus{\lbrace i \rbrace}} z_{i}\cdot z_{j} \\
    &\overset{(Q5)}{\geq} - \vert B_{y} \vert \times \frac{1}{\vert B_{y}\vert (\vert B_{y}\vert -1)} \sum_{i \in {B}_{y}}
    \sum_{j\in B_{y}\setminus{\lbrace i \rbrace}} \Vert z_{i}\Vert \Vert z_{j} \Vert = - \vert B_{y} \vert
\end{aligned}
\end{equation}
Since the $z_{i}$ and $z_{j}$ are on a hypersphere, this implies the condition of equality is equivalent to $z_{i}=z_{j}$.
\begin{enumerate}[(Q5)]
    \item For every $n,m \in [N]$, $y_{n}=y_{m}$ implies $z_{n}=z_{m}$.
\end{enumerate}
Note that (Q5) implies the variability collapse, that is all the within-class representations collapse to their class means. When this condition holds and recall the definition of balanced contrastive loss, for an instance $x_i$ with label $y$ in a batch $B$, balanced contrastive loss has the following expression:
\begin{equation}\label{eq:13}
\begin{aligned}
     \mathcal{L}_{BCL} &= \sum_{B \in \mathcal{B}} \sum_{y \in \mathcal{Y}} \sum_{i \in B_{y}} \mathcal{L}_{i} \\
     \mathcal{L}_{i}&=-\log\frac{\exp(z_{i}\cdot z_{c_{y}})}{\exp(z_{i}\cdot z_{c_{y}})+\sum \limits_{j\in \mathcal{Y}\setminus{\lbrace y \rbrace}}\exp(z_{i}\cdot z_{c_{j}})}
\end{aligned}
\end{equation}
Note that under the condition of (Q5), for every $B \in \mathcal{B}$, every $y \in \mathcal{Y}$ and every $i \in B_{y}$, it holds that $z_{i}=z_{c_{y}}$, and the label configuration of $\mathcal{L}_{i}$ is balanced.
To minimize the above loss, the solution obviously satisfies the simplex configuration. Leveraging the lower bound of supervised contrastive loss under balanced settings~\cite{dissecting}, we have
\begin{equation}
    \mathcal{L}_{i} \overset{(Q6)}{\geq} \log \left( 1+ (K-1)\exp \left(-
    \frac{K}{K-1}
    \right)
    \right)
\end{equation}
\begin{enumerate}[(Q6)]
    \item $z_{c_{1}},\dots,z_{c{_{K}}} \ form\  a\  regular\ simplex$
\end{enumerate}
Combine the aforementioned conditions, we can obtain the claimed lower bound of balanced contrastive loss:
\begin{equation}
\label{eq:lower_bound_bcl}
    \mathcal{L}_{BCL}(Z;Y) \geq \vert \mathcal{D}\vert\log \left( 1+ (K-1)\exp \left(-
    \frac{K}{K-1}
    \right)
    \right)
\end{equation}
Recall that $Z$ is an N point configuration with labels $Y$, the equality of Eq.~\ref{eq:lower_bound_bcl} is attained if and only if the following conditions hold. There are $\zeta_{1},\dots,\zeta_{K} \in \mathbb{R}^{h}$ such that:
\begin{enumerate}[(1)]
    \item $\forall n \in [N]:  z_{n}=\zeta_{y_{n}}$
    \item $\zeta_{1},\dots,\zeta_{K} \ form\  a\  regular\ simplex$
\end{enumerate}
\section{Gradient Analysis}
Balanced contrastive loss achieves the balance by averaging the parts of each class. An analysis of the gradients well reflects this conclusion. First, we will discus the defects of the supervised contrastive loss when working on the long-tailed data. Next, we will give the gradient derivation of the balanced contrastive loss, from which we can easily identify that balanced contrastive loss is better at handling long-tailed data.

Recall the definitions of supervised contrastive (SC) loss, neglecting the hyper parameter temperature $\tau$, the gradient of SC loss has the following formulation~\cite{scl}:
\begin{equation}
    \frac{\partial \mathcal{L}_{i}^{SC}}{\partial z_{i}} = \underbrace{\sum \limits_{p\in B_{y}\setminus{\lbrace i \rbrace}} z_{p}  \left(
    P_{ip} - \frac{1}{\vert B_{y} \vert -1}
    \right)}_{positive\ term} + \underbrace{\sum \limits_{n\in B_{y}^{C}} z_{n}  
    P_{in} }_{negative\ term}
\end{equation}
where $B_{y}^{C}$ is the complement set of $B_{y}$ and we have defined:
\begin{equation}
\begin{aligned}
    P_{ip}&=\frac{\exp (z_{i} \cdot z_{p})}{\sum\limits_{k\in B\setminus\{i\}}\exp(z_{i}\cdot z_{k})} \\
    P_{in}&=\frac{\exp (z_{i} \cdot z_{n})}{\sum\limits_{k\in B\setminus\{i\}}\exp(z_{i}\cdot z_{k})}
\end{aligned}
\end{equation}
Since there is a normalization function before computing the loss. Let $w_{i}$ denote the output prior to normalization in a slight abuse of notation, i.e., $z_{i}=w_{i}/\Vert w_{i} \Vert$. Then, the gradient with respect to $w_{i}$ is as follows:
\begin{equation}\label{eq:scl_grad}
\begin{aligned}
    \frac{\partial \mathcal{L}_{i}^{SC}}{\partial w_{i}} &= 
    \frac{1}{\Vert w_{i} \Vert} (I-z_{i}z_{i}^{T}) \left(
    \sum \limits_{p\in B_{y}\setminus{\lbrace i \rbrace}} z_{p}  \left(
    P_{ip} - \frac{1}{\vert B_{y} \vert -1}
    \right) + \sum \limits_{n\in B_{y}^{C}} z_{n}  
    P_{in}
    \right)\\
    &= \frac{1}{\Vert w_{i} \Vert} \left(\underbrace{
    \sum \limits_{p\in B_{y}\setminus{\lbrace i \rbrace}} (z_{p}-(z_{i} \cdot z_{p})z_{i})(P_{ip}-\frac{1}{\vert B_{y} \vert -1})}_{positive\ term} + \underbrace{
    \sum \limits_{n\in B_{y}^{C}} (z_{n} - (z_{i} \cdot z_{n})z_{i})
    P_{in}}_{negative\ term}
    \right)
\end{aligned}
\end{equation}
We mainly concern with the gradients from the negative term. For hard negatives, $z_{i}\cdot z_{n} \approx 0$ (assume $z_{i}\cdot z_{n}\leq 0$ ), so that the gradient of $\mathcal{L}_{i}^{SC}$ from the hard negatives is as follows:
\begin{equation}
\begin{aligned}
    &\sum \limits_{n\in B_{y}^{C}} \Vert z_{n} - (z_{i} \cdot z_{n})z_{i} \Vert \vert P_{in} \vert \\
    &\approx \sum \limits_{n\in B_{y}^{C}} \vert P_{in} \vert \\
    &=\sum \limits_{n\in B_{y}^{C}} \frac{1}{\sum\limits_{k\in B\setminus\{i\}}\exp(z_{i}\cdot z_{k})}
\end{aligned}
\end{equation}
Given an anchor, the term in the denominator is consistent for all negative samples, resulting in the negative class gradient is proportional to the number of samples. But under the long-tailed distribution, within almost every mini-batch, there are much more head class samples than tail class samples. This leads to all classes being as far away from the head category as possible, and results in an unbalanced feature space.

For balanced contrastive (BC) loss, the gradient has the following formulation:
\begin{equation}
\begin{aligned}
    \frac{\partial \mathcal{L}_{i}^{BC}}{\partial z_{i}} &= - \frac{1}{\vert B_{y} \vert -1} \sum \limits_{p\in B_{y}\setminus\{i\}} \left(
    z_{p} - \sum \limits_{j\in \mathcal{Y}} \frac{1}{\vert B_{j}\vert} 
    \sum \limits_{k\in B_{j}} z_{k} X_{ik}
    \right) \\
    &= - \frac{1}{\vert B_{y}\vert -1} \sum \limits_{p\in B_{y}\setminus\{i\}} \left(
    z_{p} - \frac{1}{\vert B_{y}\vert -1} \sum \limits_{p'\in B_{y}\setminus\{i\}} z_{p'} X_{ip'}
    -
    \sum \limits_{j\in \mathcal{Y}\setminus\{y\}} \frac{1}{\vert B_{j}\vert}
    \sum \limits_{k\in B_{j}} z_{k} X_{ik}
    \right) \\
    &= \underbrace{\frac{1}{\vert B_{y}\vert -1} \sum \limits_{p\in B_{y}\setminus\{i\}}
    z_{p} (X_{ip}-1)}_{positive\ term} + \underbrace{
    \sum \limits_{j\in \mathcal{Y}\setminus\{y\}} \frac{1}{\vert B_{j}\vert}
    \sum \limits_{k\in B_{j}} z_{k} X_{ik}}_{negative\ term}
\end{aligned}
\end{equation}
where we have defined:
\begin{equation}
\begin{aligned}
    X_{ip} &= \frac{\exp (z_{i} \cdot z_{p})}{\sum \limits_{j\in \mathcal{Y}} \frac{1}{\vert B_{j} } \vert
    \sum \limits_{k\in B_{j}}\exp (z_{i} \cdot z_{k})} \\
    X_{ik} &= \frac{\exp (z_{i} \cdot z_{k})}{\sum \limits_{j\in \mathcal{Y}} \frac{1}{\vert B_{j} } \vert
    \sum \limits_{k\in B_{j}}\exp (z_{i} \cdot z_{k})}
\end{aligned}
\end{equation}
Similar to the derivation of supervised contrastive loss, the gradient with respect to $w_{i}$ of balanced contrastive loss is as follows:
\begin{equation}\label{eq:bcl_grad}
\begin{aligned}
    \frac{\partial \mathcal{L}_{i}^{BC}}{\partial w_{i}} = \frac{1}{\Vert w_{i} \Vert} \left(
    \underbrace{\frac{1}{\vert B_{y}\vert -1} \sum \limits_{p\in B_{y}\setminus\{i\}}
    (z_{p}-(z_{i} \cdot z_{p})z_{i}) (X_{ip}-1)}_{positive\ term} + \underbrace{
    \sum \limits_{j\in \mathcal{Y}\setminus\{y\}} \frac{1}{\vert B_{j}\vert}
    \sum \limits_{k\in B_{j}} (z_{k}-(z_{i} \cdot z_{k})z_{i}) X_{ik}}_{negative\ term}
    \right)
\end{aligned}
\end{equation}
Intuitively, balanced contrastive loss balances the gradients from negative classes, avoiding a tremendous gradient update from the negative head class samples. It retains several good properties of supervised contrastive loss. Easy negatives $z_{i}\cdot z_{k} \approx -1$ contributes less gradient while hard negatives more gradient, and easy positives $z_{i}\cdot z_{p} \approx 1$ (assume $z_{i}\cdot z_{p} \geq 0$), contributes less gradient compared with hard positives. In addition to these common properties, the balanced contrastive loss is better at feature alignment, where points belonging to the same class are pulled together.
Since almost every mini-batch is long-tailed, for these head class anchors, the gradients in Eq.~\ref{eq:scl_grad} from the positives will be much larger than when the anchor is tails. It results in tail class samples being unconcerned to pulling these points together.
Comparing Eq.~\ref{eq:scl_grad} with Eq.~\ref{eq:bcl_grad}, balanced contrastive loss also adjusts the gradients from the positives, eliminating excessive gradient fluctuations caused by having different anchor classes in different batches and allowing the points of tail classes been pulled closer.

\section{More Results}
\subsection{Ablations of Different Forms of Prototypes.}
We compare our method with the other two implementations of the prototype. The first one is using the exponential moving average to calculate the prototype. The second one is using learnable parameters~\cite{paco, hybrid}. As shown in Table~\ref{tab:prototype}, our implementation achieves the best results and the other two implementations achieve similar results.  
\begin{table}[h]
    \centering
    \begin{tabular}{c c c c c}
    \hline
         Methods & Many & Medium & Few & All  \\
         \hline
        Exponential Moving Average & 69.7 & 54.4 & 31.9 & 53.0\\
        Learnable Parameters & 68.9 & 54.0 & 34.3& 53.3\\
        Ours & 69.7 & 53.8 & 35.5 & 53.9\\
        \hline
\end{tabular}
    \caption{Ablation study for different implementations of prototypes on CIFAR-100-LT with an imbalance fator of 100. All models run for 400 epochs with the same training scheme.}
    \label{tab:prototype}
\end{table}
\subsection{Ablations of Different Configurations of Views.}
We compare our configuration with the other two configurations of views. We use the simple augmentation method, i.e., SimAug, to generate both views for contrastive learning. We further have one of the views generated via a stronger augmentation method, i.e., RandAug, or both of the views generated by RandAug. As shown in Table~\ref{tab:view}, stronger argumentation yields better performance.
\begin{table}[h]
    \centering
    \begin{tabular}{c c c c c}
    \hline
         Methods & Many & Medium & Few & All  \\
         \hline
        SimAug.$\&$SimAug. & 67.2 & 53.9 & 36.5 & 56.7\\
        RandAug.$\&$SimAug. & 67.1 & 54.6 & 37.1 & 57.1\\
        RandAug.$\&$RandAug. & 67.6 & 54.6 & 37.5 & 57.3\\
        \hline
\end{tabular}
    \caption{Ablation study for different configurations of views on ImageNet-LT. All models run for 90 epochs with the same training scheme.}
    \label{tab:view}
\end{table}
\subsection{Confusion Matrix.}
\begin{figure*}[h] 
  \centering
  \vspace{-10pt}
  \begin{subfigure}[]{0.33\linewidth}
    \includegraphics[scale=0.4]{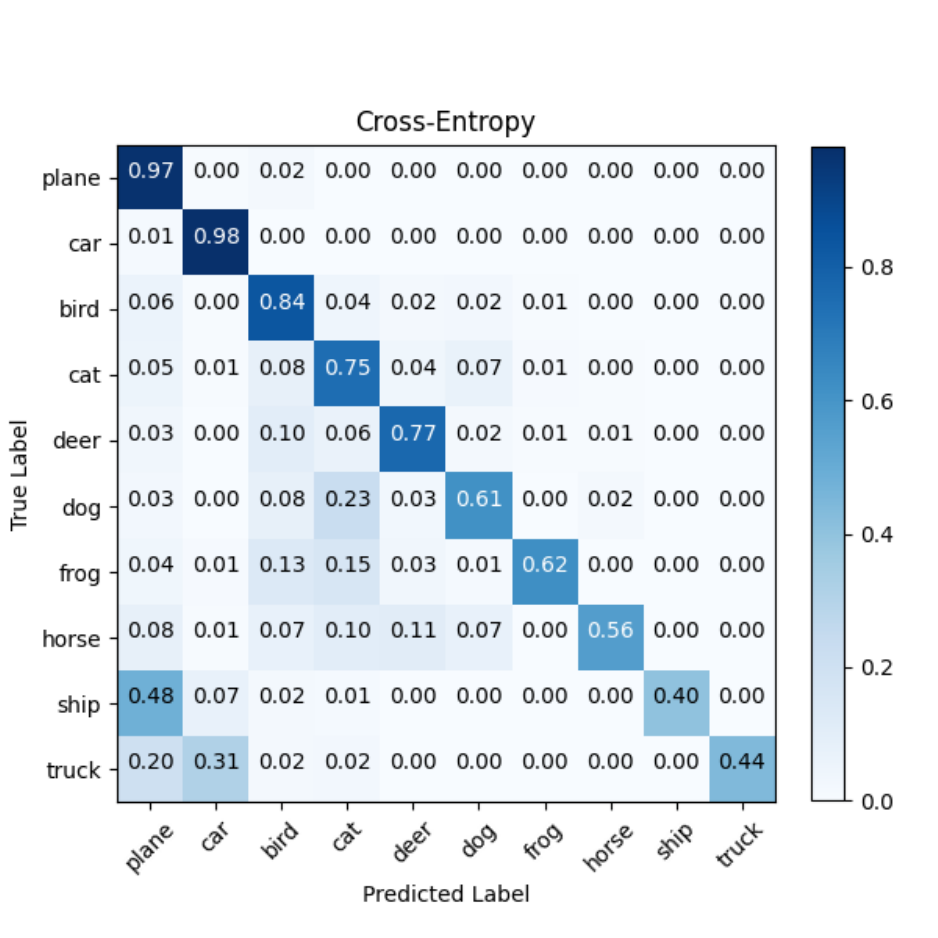}
    \caption{}
    \label{fig:cf1} 
  \end{subfigure}%%
  \centering
  \begin{subfigure}[]{0.33\linewidth}
    \includegraphics[scale=0.4]{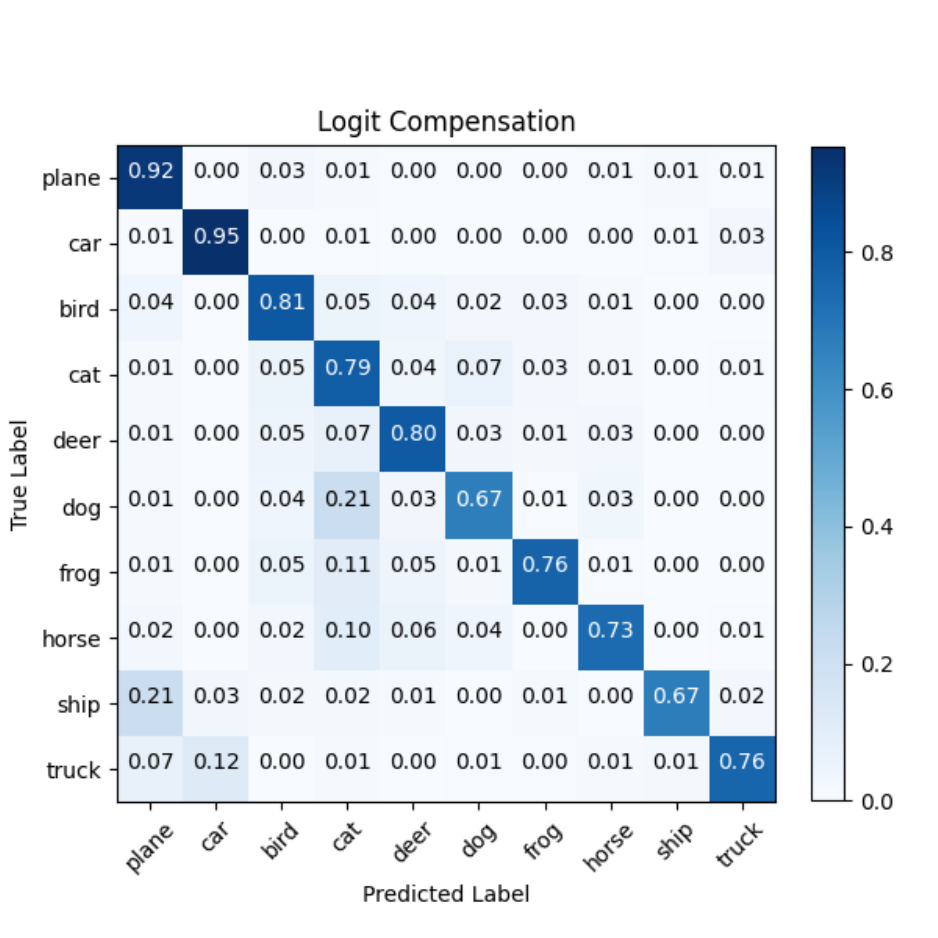} 
    \caption{}
    \label{fig:cf2} 
  \end{subfigure}%%
  \centering
  \begin{subfigure}[]{0.33\linewidth}
    \includegraphics[scale=0.4]{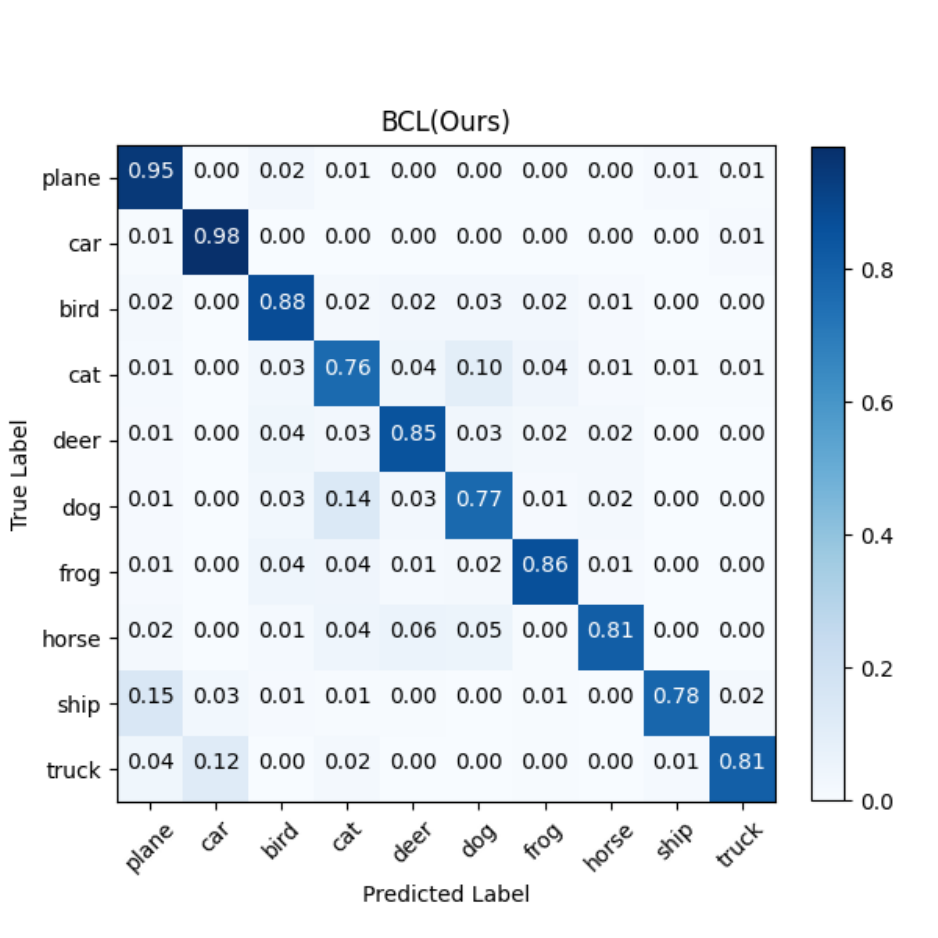} 
    \caption{}
    \label{fig:cf3}
  \end{subfigure}
  \caption{Illustration of confusion matrix of prediction results on CIFAR-10-LT for different models.
    }
    \label{fig:cf}
\end{figure*}

To clearly show where the models are getting confused on long-tailed data, we illustrate the confusion matrix of predictions on CIFAR-10-LT in Figure~\ref{fig:cf}. With vanilla cross-entropy, the model tends to misclassify low-frequency artifactory categories as high-frequency artifactory categories and low-frequency animal classes as high-frequency animal classes. With logit compensation, misclassification of low-frequency classes is greatly eased. With the proposed BCL, low-frequency classes are more correctly predicted than high-frequency classes, and the accuracies of high-frequency classes are also improved.

% \begin{figure*}[t]
% \centering
% \subfigure[]{\begin{minipage}[h]{0.33\linewidth}
% \includegraphics[scale=.33]{fig/confusion1.pdf}
% \end{minipage}
% }%
% % \subfigure[]{\label{fig:3a}%
% % \includegraphics[scale=.33]{fig/confusion2.pdf}}%
% % \subfigure[]{\label{fig:3b}%
% % \includegraphics[scale=.33]{fig/confusion3.pdf}}%
% % \caption{Maximum rank ratio per-layer of ResNet-18.}
% \end{figure*}

\subsection{Visualization of Learned Features.}
\begin{figure*}[h] 
  \centering
  \begin{subfigure}[]{0.5\linewidth}
    \includegraphics[scale=0.5]{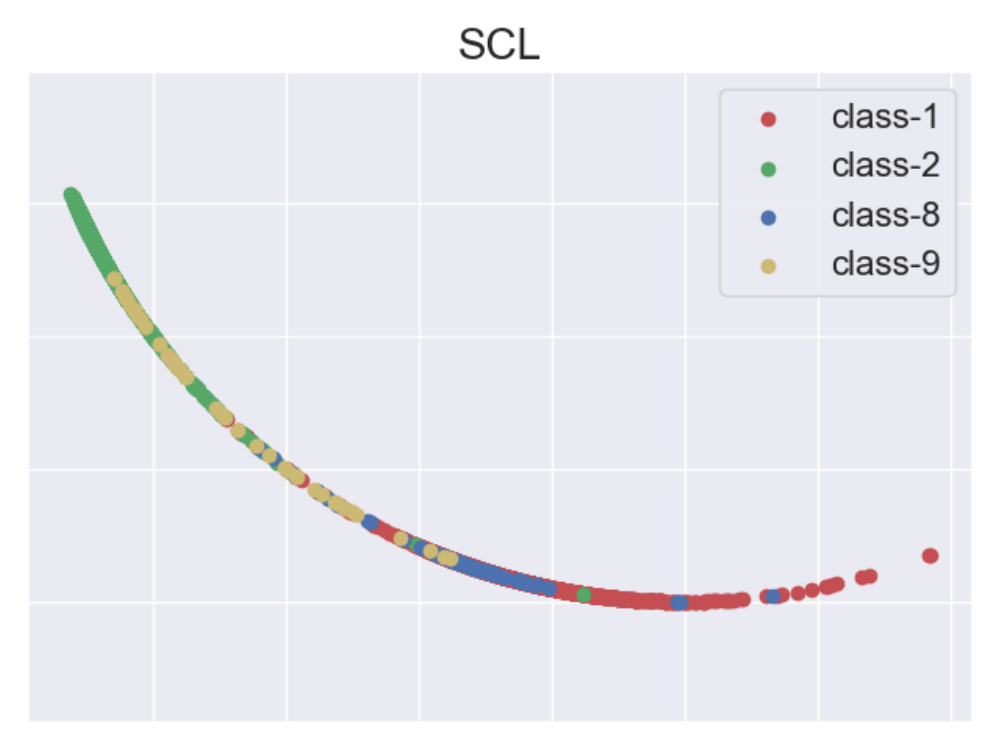}
    \caption{}
    \label{fig:scl} 
  \end{subfigure}%%
  \centering
  \begin{subfigure}[]{0.5\linewidth}
    \includegraphics[scale=0.5]{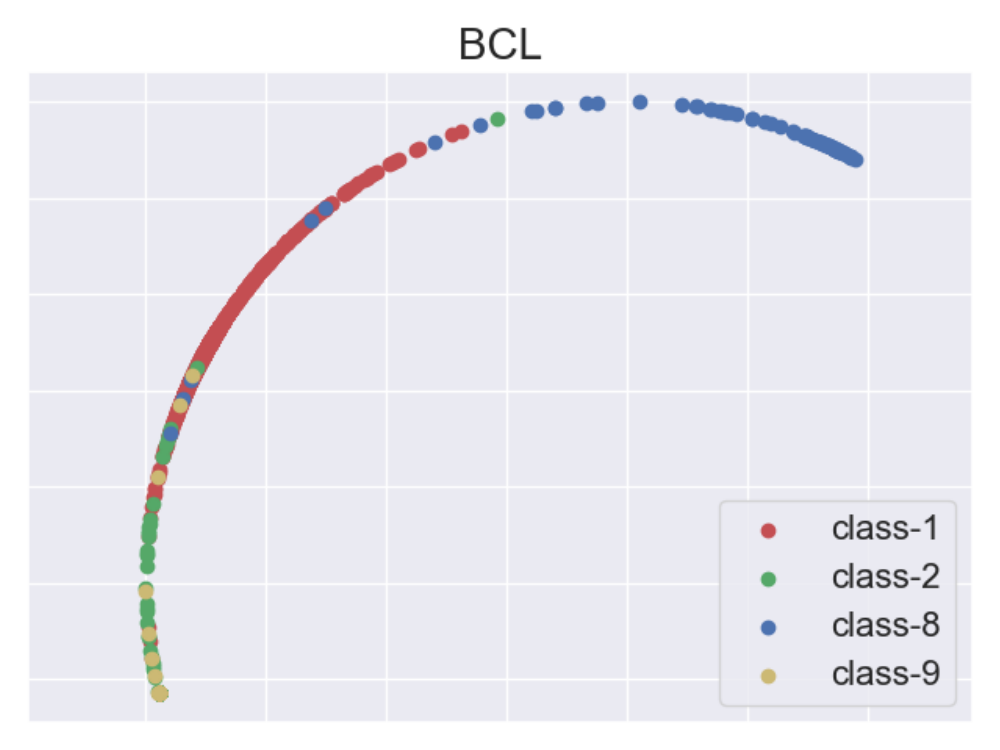} 
    \caption{}
    \label{fig:bcl} 
  \end{subfigure}%%
  \caption{Illustration of features learned by SCL and BCL.
    }
\end{figure*}

Similar to~\cite{li2021targeted}, we visualize the 2-dimensional MLP output feature learned by SCL and BCL on CIFAR-10-LT. 
Features of different classes learned by BCL distribute more uniform on the sphere and are more separable than SCL. 

\end{document}